\documentclass[11pt]{article}

\usepackage[final]{acl}

\usepackage{times}
\usepackage{latexsym}

\usepackage[T1]{fontenc}

\usepackage[utf8]{inputenc}

\usepackage{microtype}

\usepackage{inconsolata}

\usepackage{graphicx}

\usepackage[utf8]{inputenc} 
\usepackage[T1]{fontenc}    
\usepackage{hyperref}       
\usepackage{url}            
\usepackage{booktabs}       
\usepackage{amsfonts}       
\usepackage{nicefrac}       
\usepackage{microtype}      
\usepackage{xcolor}         
\usepackage{graphicx} 
\usepackage{array}
\usepackage{multirow} 
\usepackage{amsmath}
\usepackage{amssymb}
\usepackage{colortbl}
\definecolor{lightgreen}{RGB}{226, 239, 218}
\definecolor{lightblue}{RGB}{68, 114, 196}
\definecolor{lightyellow}{RGB}{255, 233, 158}
\definecolor{limegreen}{RGB}{90, 111, 62}
\usepackage{wrapfig}
\usepackage{enumitem}

\usepackage{arydshln}
\usepackage{CJK} 
\usepackage{placeins}

\usepackage{inconsolata}
\usepackage[linesnumbered,ruled]{algorithm2e}
\usepackage{algorithmic}
\usepackage{amsmath,amsfonts,amssymb}
\usepackage{mathtools}

\title{PlanE: Meta Planning of Data, Tuning, and Inference for\\ Extractive-based LLMs}

\author{
 \textbf{Jiacheng Wang$^{\diamondsuit}$\thanks{~Equal Contribution.}},
 \textbf{Weiyan Zhang$^{\spadesuit}$\footnotemark[1]}\thanks{~Corresponding Authors.},
 \textbf{Guangya Yu$^{\spadesuit}$}
 \\
 $^{\diamondsuit}$Ant Group, Shanghai, China,
\\
 $^{\spadesuit}$School of Information Science and Engineering, East China University\\of Science and Technology, Shanghai, China
 \\
 \small{
   \textbf{Correspondence:} \href{xueqiao.wjc@antgroup.com}{xueqiao.wjc@antgroup.com},
   \href{weiyanzhang@ecust.edu.cn}{weiyanzhang@ecust.edu.cn}
 }
}

\begin{document}
\maketitle
\begin{abstract}
Enhancing the task-specific capabilities of Large Language Models (LLMs) primarily requires substantial instruction-tuning datasets. However, the sheer volume of such data imposes a considerable annotation cost, and a lack of optimization methods for tailoring LLMs to specific tasks. To address the above issues, we propose a \textbf{Plan}ning framework for constructing \textbf{E}xtractive-based LLMs called \textbf{PlanE}, which includes data decomposition, instruction tuning, and prompt inference. Additionally, we introduce a Data-Tuning-Inference (DTI) planner, aimed at selecting the optimal base-LLM and its DTI combinations for specific datasets to improve construction efficiency. The experimental results demonstrate the effectiveness of our PlanE from two views: (1) across different datasets using the same base-LLM, and (2) on the same dataset using different base-LLMs. Furthermore, we validate the generalizability of the proposed DTI planner under different optimization objectives. The codes are publicly available at \url{https://github.com/gugugu-469/PlanE}.
\end{abstract}


\begin{figure}[!bt]    
\centering
\includegraphics[width=0.43\textwidth]{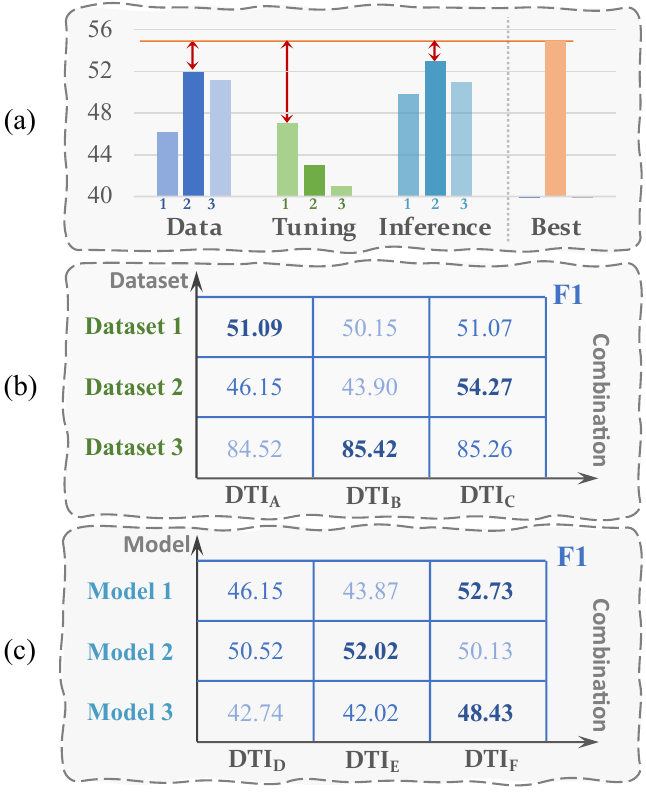}
\vspace{-0.2cm}
\caption{(a) Three key factors in building task-specific LLMs.
(b) For the same base-LLM, the optimal performance differs across different task datasets.
(c) On the same dataset, different base-LLMs reach their best performance through different combinations of DTI.}
\vspace{-0.4cm}
\label{fig:intro}
\end{figure}



\section{Introduction}
LLMs have demonstrated prowess in producing human-aligned responses to varied instructions. A pivotal technique for enhancing the task-specific capabilities of LLMs is instruction tuning, which aligns the model with human preferences using data in the form of instruction-response pairs.

Based on the stage of LLM optimization, these works can be categorized into three approaches. First, data refinement employs data augmentation and data structure transformation to improve LLM performance~\cite{dingetal2024data,chang2024data}. However, these methods are often constrained by data quality~\cite{lietal2024empowering}. Second, multi-task tuning trains LLMs on multiple information extraction (IE) tasks simultaneously to improve generalization~\cite{abdelazizetal2024granite}. However, such methods do not significantly boost the performance of the specific task and come with high tuning costs~\cite{brief2024mixing}. Third, retrieval-augmented generation integrates retrieval mechanisms with LLMs to enhance task performance through external knowledge bases (KBs)~\cite{wangetal2025llms,suetal2024dragin}. However, this approach is limited by the challenges associated with building specialized KBs~\cite{lietal2024retrieval}.





Current research primarily focuses on the use of LLM technology, but there is a lack of studies on constructing the LLM for IE-specific tasks. When building an IE-specific LLM, it is necessary to consider the optimization of the entire process, including data, tuning, and inference, as shown in Figure~\ref{fig:intro} (a): 1) \textbf{Data}. Given the limited training data, constructing high-quality instruction-tuning data is a significant challenge. 2) \textbf{Tuning}. For training data with different data structures, determining the best way to combine tuning strategies to maximize the potential of LLMs is key. 3) \textbf{Inference}. For a tuned LLM, designing prompts to fully leverage its reasoning capabilities remains another critical issue.
Furthermore, the integration and joint optimization of these three stages into a comprehensive solution is also a major difficulty. This is because, as shown in Figure~\ref{fig:intro} (b), even when using the same three Data-Tuning-Inference (DTI) combinations, the optimal performance varies across different task datasets for the same base-LLM. Similarly, as shown in Figure~\ref{fig:intro} (c), even on the same dataset, different base-LLMs achieve their best performance with different DTI combinations, despite using the same three DTI combinations.








In this paper, we propose a novel framework for selecting combinations to construct extractive-based LLMs, called \textbf{PlanE}. 
The core idea is to design optimization schemes for the three key stages of the extractive-based LLM construction process (i.e., data, tuning, inference), and propose a combination selection function to determine the optimal construction strategy. 
Specifically, for the data, we introduce two data decompositions: Pipeline-based and Bidirectional. These schemes decompose the complex task into a combination of serialized atomic subtasks. Then, for tuning, we adopt instruction tuning. Based on the decomposed data structure of subtasks, we employ Supervised Fine-Tuning (SFT), and a combination of SFT with Reinforcement Learning (SFT+RL).
Finally, for inference, we adopt prompt inference, including three inference strategies: direct inference, intersection inference, and union inference.
In addition, we design a DTI planner that predicts the optimal DTI combination for given task datasets and base-LLMs through meta-learning on empirical data, thereby optimizing extractive-based LLM construction.

The contributions of this paper are as follows:
\begin{itemize}
    \item We design a novel DTI planner that selects the optimal DTI combinations for the extractive-based LLM. Besides, this planner also performs multi-objective optimization, balancing both the performance and efficiency of building the extractive-based LLM.
     \item We propose the PlanE framework, which enhances high-performance LLM construction through three key modules. From a holistic optimization perspective, it determines the appropriate DTI combination based on the specified base-LLM and dataset.
    

    \item We conduct extensive experiments on three task datasets. The results indicate the effectiveness of our approach from two views: across different task datasets using the same base-LLM, and on the same task dataset using different base-LLMs. In addition, we validate the generalizability of the proposed DTI planner under different optimization objectives.
\end{itemize}

\begin{figure*}[!bt]     
\centering
\includegraphics[width=0.95\textwidth]{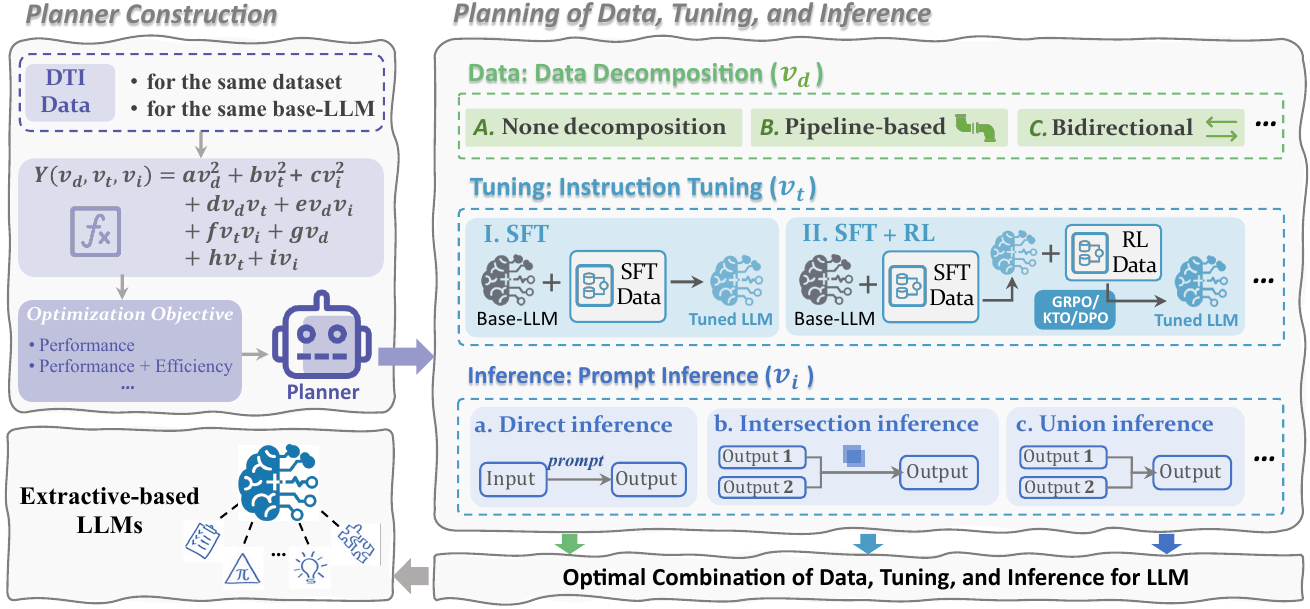}
\vspace{-0.2cm}
\caption{The framework of PlanE.}
\vspace{-0.4cm}
\label{Fig.2}
\end{figure*}

\section{Related Work}\label{2}
Recent work in meta-planning for task-specific LLMs can be broadly categorized into several approaches: 1) policy space design, 2) meta-policy learning, and 3) optimization objectives.

The first branch is policy space design, which defines the set of high-level strategies or meta-actions available to a meta-planner. For instance, DeepSpeed-Inference~\cite{aminabadi2022deepspeed} defines its policy space through kernel selection and parallelism configurations. DyPlan~\cite{parekh2025dynamic} constructs a dynamic policy space by composing reasoning paths (e.g., Chain-of-Thought, Retrieval-Augmented Generation). 
In decentralized settings, MetaInf~\cite{du2025metalearning} leverages semantic embeddings and historical metadata to predict acceleration strategies, enabling zero-shot generalization without online profiling. 

The second branch, meta-policy learning, refers to constructing or learning the optimal mapping from context to actions within the policy space. While MetaGPT~\cite{hong2024metagpt} relies on handcrafted workflows, PromptAgent~\cite{wang2024promptagent} employs reinforcement learning to discover dynamic reasoning and prompting strategies, effectively learning a meta-policy. Similarly, data-level planning methods like DataRater~\cite{calian2025datarater} and GradMatch~\cite{killamsetty2021grad} for sample selection, and MoE-Scaling~\cite{tian2025towards} for expert configuration, can be viewed as specialized forms of meta-policy learning for data and architecture management.

The third branch, optimization objectives, defines the criteria or metrics that guide the selection and evaluation of policies within the policy space, such as latency, accuracy, cost, or efficiency. DeepSpeed-Inference balances latency and accuracy, while AdaCoT~\cite{lou2025adacot} uses Pareto search to navigate cost-performance trade-offs. Meta-R1~\cite{dong2025meta} introduces root-scaled efficiency for cross-scale efficiency normalization, and Chinchilla~\cite{Chinchilla} provides a compute-optimal scaling law that has been adopted in Llama 2~\cite{touvron2023llama}.

Existing policy spaces are often limited to isolated aspects, lacking a unified representation that jointly covers data, training, and inference. 
Meta-policy learning remains fragmented, with some methods relying on handcrafted rules and others optimizing only local decisions. 
Optimization objectives are either narrowly decoupled or based on idealized scaling laws that overlook practical multi-objective trade-offs.
To address these issues, this paper introduces a data-driven planner that unifies the full configuration space into a single policy representation, learns an integrated meta-policy via empirical performance modeling, and directly optimizes joint performance–efficiency objectives.

\section{Overview}\label{3}
In this section, we first formally define three IE tasks. Then, we outline our proposed PlanE.


\subsection{Task Formulation}\label{3.1}
\textbf{Definition 1: Relation extraction task (RE)} Given an input text $C=[c_1,c_2, ..., c_n]$ and a set of pre-defined relations $\mathcal{R}=\lbrace r_1,r_2, ..., r_l\rbrace$, where $n$ and $l$ denote the numbers of tokens in $C$ and relations in $\mathcal{R}$, respectively, the purpose of relation extraction task is to obtain a set of triplets $\mathcal{T}=\lbrace(h_i,r_i,t_i)\rbrace_{i=1}^{m}$, where $h_i,\ t_i$ represent the head and tail entities connected by $r_i$ and $m$ denotes the total number of extracted triplets.


\noindent\textbf{Definition 2: Event extraction task (RE)} Given a text $C_{e}=[c_1,c_2, ..., c_{n_e}]$ and the pre-defined event types $\mathcal{E}=\lbrace e_1,e_2, ..., e_{l_e}\rbrace$, event extraction task aims to detect the mention event trigger $tr$ and arguments $\mathcal{A}=\lbrace a_1,a_2, ..., a_{m_e}\rbrace$ from $C_{e}$ and the event type $e \in \mathcal{E}$ (e.g., \texttt{EAT}, \texttt{LOCATION}, etc).

\noindent\textbf{Definition 3: Aspect-based sentiment analysis task (ABSA)} Given a text $C_{a}=[c_1, c_2, ..., c_{n_a}]$, a set of pre-defined sentiment polarities $\mathcal{S}=\{s_1, s_2, ..., s_{l_a}\}$ (e.g., $\lbrace\texttt{POSITIVE}, \texttt{NEGATIVE}, \texttt{NEUTRAL}\rbrace$), where $n_a$ and $l_a$ is the number of tokens in $C_{a}$ sentiment polarities in $\mathcal{S}$, respectively, the ABSA task aims to extract a set of aspect-sentiment triplets $\mathcal{Y}=\{(at_j, o_j, s_j)\}_{j=1}^{k}$ from $C_{a}$. $k$ is the number of aspect-sentiment triplets and $at_i$, $o_i$ denote the aspect term and opinion term in $C_{a}$.

\subsection{Framework}\label{3.2}
As shown in Figure~\ref{Fig.2}, our solution is to construct an extractive-based LLM (e.g., RE-LLM) in two parts. 
\textit{1) Planning of Data, Tuning, and Inference.} Specifically, for the data, given the training set, we introduce two Data Decompositions: Pipeline-based and Bidirectional. Then, for tuning, based on the decomposed data structure of subtasks, we employ two Instruction Tuning: SFT and SFT+RL. 
Finally, for inference, based on tuned LLMs, we adopt three Prompt Inference strategies: direct inference, intersection inference, and union inference.
\textit{2) Planner Construction.} To achieve automatic prediction of the optimal DTI combination, we design a DTI Planner. This planner predicts the optimal DTI combination for given datasets and base-LLMs through meta-learning on empirical data, thereby optimizing extractive-based LLM construction.

\section{PlanE}\label{4}
This section details PlanE, including: (1) Planning of Data (\textit{data decomposition}), Tuning (\textit{instruction tuning}), and Inference (\textit{prompt inference}), (2) Planner Construction, and (3) Planner Usage. 


\subsection{Planning of Data, Tuning, and Inference}\label{4.1}

\textbf{Data Decomposition.}
This section presents a task decomposition method aimed at enhancing the performance of language models on complex tasks through data augmentation. By decomposing a task into subtasks, we improve the precision of LLMs at each step and reduce error propagation. We propose two decomposition strategies: pipeline-based and bidirectional. In pipeline-based decomposition, the task is broken down into sequential subtasks, where the output of one subtask serves as the input for the next. For example, in RE task, this can be implemented in two ways: first, performing entity recognition (ER) to extract head-tail entity pairs, followed by relation classification (RC) to generate triplets; second, performing RC to determine the relation, then ER to extract the corresponding head-tail entity pairs and output triplets. In bidirectional decomposition, the task is approached from two directions to improve precision. For example, in RE, first extracting the head entity, generating the relation, and then extracting the tail entity, or starting with the tail entity and proceeding accordingly.

\textbf{Instruction Tuning.}
To align the model with diverse task structures and subtask objectives, we employ two tuning strategies: 1) SFT. All data is formatted into structured input-output instruction pairs, and a single model is fine-tuned to handle the RE task and its subtasks.
2) SFT + RL. It is a two-stage instruction tuning strategy. First, the model is trained with SFT using the majority of the data, where all samples are formatted into structured input-output instruction pairs to enable general task-solving capability. Then, RL is applied on the remaining subset to further optimize model performance through feedback-based reward mechanisms.
Specifically, we adopt three RL-based optimization methods, namely Group Relative Policy Optimization (GRPO)~\cite{shao2024deepseekmath}, Direct Preference Optimization (DPO)~\cite{rafailov2023direct}, and Kahneman-Tversky Optimization (KTO)~\cite{ethayarajh2024kto}. Detailed information is listed in Appendix~\ref{app:H}.

\textbf{Prompt Inference.}
To leverage the strengths of different decomposition strategies during inference, we propose the following methods: 1) Direct Inference. The model directly processes the input and generates the final output; 2) Intersection Inference. Aggregates results by retaining only outputs consistent across different inference chains. 3) Union Inference. Combines results from multiple inference chains (e.g., pipeline-based or bidirectional) by taking their union. For example, the RE task is based on pipeline-based data decomposition. For direct inference,  we directly output relation triplets from the given text. For union/intersection inference, we first perform ER to extract head-tail entity pairs, followed by RC to generate triplets, then perform RC to determine the relation, followed by ER to extract entity pairs. Finally, we aggregate results at the triplet level using union or intersection. 
Detailed prompts are listed in Appendix~\ref{app:G}.


\subsection{Planner Construction}\label{4.2}

The construction of the DTI planner for various objectives (e.g., improve performance) consists of five key steps:

1. \textbf{\textit{Planner definition:}}
We define a planner that selects the optimal plan based on the task context. The planner is trained on historical execution data, enabling it to automatically identify and select the best-performing plan in specific scenarios.  
Formally, given the historical execution data,
\begin{equation}
\mathcal{D} = \{ (t_k^{(i)}, a^{(i)}, r^{(i)}) \}_{i=1}^N,
\end{equation}
where \(t_k^{(i)} \in \mathcal{T}\) denotes the $k$-th task, \(a^{(i)} \in \mathcal{A}\) is the selected plan, and \(r^{(i)} \in \mathbb{R}\) represents the performance feedback (e.g., the F1-score, which provides a more comprehensive measure for selecting the best combination as the harmonic mean of precision and recall) of the plan on task \(t_k^{(i)}\).  
The planner \(\mathcal{P}_x\) is implemented by a parameterized ranking model \(f_x: \mathcal{T} \times \mathcal{A} \rightarrow \mathbb{R}\), which satisfies,  
\begin{equation}
\mathcal{P}_x(t) = \arg\max_{a \in \mathcal{A}} f_x(t, a),
\end{equation} 
meaning that it selects the plan with the highest F1 among all candidates. The scoring function \(f_x\) is trained on $\mathcal{D}$ using a learning-to-max objective.

2. \textbf{\textit{Element definition:}} In this paper, we explore two scenarios: planners across different datasets and planners across different models. Accordingly, \(t_{k}^{(i)}\) represents either a model or a dataset. \(a^{(i)}\) denotes different DTI combinations, 
with a total of $Z$ candidate DTI combinations $CO$. \(r^{(i)}\) represents the performance metric (e.g. F1) achieved by the model under \(t_{k}^{(i)}\) and \(a^{(i)}\).

3. \textbf{\textit{Identify optimal combination:}} We identify the combination that yields the highest F1. This ensures that subsequent optimization is grounded in the best observed result. Specifically, we determine the optimal combination $co_{max}$ as:
\begin{equation}
co_{max} = \arg\max_{1 \le j \le num(CO)} \; (F1_j).
\end{equation}

4. \textbf{\textit{Build inequality constraints:}} To ensure that our planner correctly ranks the optimal DTI combination above all other candidates, we construct a set of inequality constraints comparing each non-optimal configuration with the optimal one.  
Besides, since each DTI combination consists of qualitative labels, we encode them into quantitative variables to enable numerical comparison and parameterized modeling within the DTI planner.  
Specifically, we define the value ranges of three integer variables (e.g., ranging from 1 to 10).
\(vd\), \(vt\), and \(vi\) represent Data Decomposition $co_d$, Instruction Tuning $co_t$, and Prompt Inference $co_i$.  
The detailed calculation methods for these variables are provided in Appendix~\ref{app:C}.  


In constructing the planner, we adopt a quadratic polynomial function with coefficients $s_{o,m}=\{a,b,c,d,e,f,g,h,i\}$ to model the nonlinear relationships between the DTI variables (i.e., \(vd\), \(vt\), and \(vt\)) and the F1.  
\(y\) denotes the predicted score of the model under a given combination, reflecting the relative effectiveness of the DTI combination:
\begin{equation}
\begin{split}
y(co) = &
a \cdot vd^2 + b \cdot vt^2 + c \cdot vi^2 + \\
&  d \cdot vd \cdot vt + e \cdot vd \cdot vi + f \cdot \\
& vt \cdot vi + g \cdot vd + h \cdot vt + i \cdot vi.
\end{split}
\end{equation}

For each non-optimal combination, the inequality constraint is formulated as:
\begin{equation}
y(vd_j, vt_j, vi_j) \leq y(vd_{max}, vt_{max}, vi_{max}).
\label{eq}
\end{equation}

These constraints ensure that the polynomial function assigns a higher score to the best-performing combination during optimization, thereby guiding the model to learn a consistent and interpretable representation of \textbf{DTI effectiveness}.

5. \textbf{\textit{Solve coefficients:}} 
For each \(t\) (representing a model or a dataset), we can obtain its corresponding set of constraints based on the observed F1 performance under different DTI configurations.
We then integrate all constraints across scenarios to jointly solve for the coefficients that define a unified planner suited for the current model or dataset.
We optimize the coefficients \(s_{o,m}=\{a,b,c,d,e,f,g,h,i\}\) of the quadratic model using gradient descent.  
We define a differentiable loss function that penalizes violations of the optimality constraints.  
Formally, the objective is written as:
\begin{equation}
\begin{split}
\mathcal{L}_{planner} = &
\min_{s_{o,m}}
\sum_{j \neq \mathrm{max}}
\mathrm{ReLU}\!\big(
y(co_j) \\
& - y(co_{max})
+ \varepsilon
\big),
\end{split}
\end{equation}
where \(\varepsilon = 10^{-10}\) is a small margin.
This gradient-based method ensures the learned polynomial ranks the optimal DTI combination higher than all others.

\subsection{Planner Usage}\label{4.3}
Once the DTI planner is obtained for a specific scenario, it can be applied to a new model or dataset.  
For example, according to the parameter calculation method (see Appendix~\ref{app:C}), only a subset of 10 representative combinations among the 28 candidates is required to infer all variable values.  
These values are then fed into the planner, and the optimal DTI combination is determined by maximizing the quadratic function:
\begin{equation}
co_{max} = \arg\max_{co \in CO}\; y(vd, vt, vi).
\end{equation}

\begin{table}[!bt]
\centering
\scalebox{0.65}{
\begin{tabular}{p{1.8cm}p{0.8cm}<{\centering}p{1.1cm}<{\centering}p{0.7cm}<{\centering}p{1.4cm}<{\centering}p{0.8cm}<{\centering}p{0.8cm}<{\centering}p{1.0cm}<{\centering}}
\toprule
\multirow{4}{*}{\textbf{Dataset}} & \multirow{4}{*}{\textbf{Lang.}} & \multirow{4}{*}{\textbf{Task}} & \multirow{4}{*}{\textbf{Dom.}} &\multicolumn{3}{c}{\textbf{\# Train}}                      & \multirow{4}{*}{\textbf{\# Test}} \\ \cmidrule(lr){5-7}
                                  &      &                               &                                     & \multirow{3}{*}{\textbf{Only SFT}} & \multicolumn{2}{c}{\textbf{SFT+RL}} &                                \\ \cmidrule(lr){6-7}
                                  &                                    &                                     &    &   & SFT      & RL     &                                \\ \midrule
CMeIE-V2                          & zh                            & RE         & Med.      & 7,000*              & 6,000*              & 1,000*            & 3,585                           \\
ACE05                             & en                            & EE        & Gen.              & 3,342              & 2,842              & 500             & 293                            \\
14Lap                             & zh                           & ABSA         & Gen.                        & 906               & 756               & 150             & 328                            \\ \bottomrule
\end{tabular}}
\vspace{-0.2cm}
\caption{Statistics of all datasets. ``\#'' = sample number, ``Lang.'' = language, ``Dom.'' = domain, ``Med.'' = medical, ``Gen.'' = general, ``SFT'' = supervised fine-tuning, and ``RL'' = reinforcement learning. The preprocessed dataset (denoted by ``*'') contains 7,000 random training samples from the original 14,339 to mitigate overfitting.}
\vspace{-0.4cm}
\label{tab:1}
\end{table}

\textbf{\textit{The optimal DTI combination usage:}}
After obtaining the optimal DTI combination, the workflow proceeds in three stages.  
First, the data decomposition $co_d$ determines how the dataset is processed for training.  
Second, the instruction tuning $co_t$ specifies the tuning strategy used to tune the LLM.  
Finally, the prompt inference $co_i$ determines the reasoning method applied during prediction.  
The tuned-LLM then produces its final outputs according to this inference strategy, completing the full DTI optimization.

\begin{table*}[!bt]
\centering
\scalebox{0.7}{
\begin{tabular}{p{2.8cm}p{3.cm}p{2.5cm}<{\centering}p{3.cm}p{2.5cm}<{\centering}p{3.cm}p{2.5cm}<{\centering}}
\toprule
\multirow{2}{*}{\textbf{Method}} & \multicolumn{2}{c}{\textbf{CMeIE-V2}}                  & \multicolumn{2}{c}{\textbf{ACE05}}                     & \multicolumn{2}{c}{\textbf{14Lap}}                                   \\ \cmidrule(lr){2-3} \cmidrule(lr){4-5} \cmidrule(lr){6-7} 
                               & Data Structure                         & Training size  & Data Structure                         & Training size  & Data Structure                              & Training size  \\ \midrule
None                   & $ c \rightarrow   (h,r,t) $            & 7,000                & $ c \rightarrow   (tr,e,a) $              & 3,342                & $ c \rightarrow   (at,o,s) $              & 906                                 \\ \cmidrule(lr){2-7} 
\multirow{5}{*}{Pipeline-based} & $ c \rightarrow (h,t)   $              & 7,000                & $ c \rightarrow (tr,a)   $                & 3,342                & $ c \rightarrow (at,o)   $                & 906                                \\
                       & $ c \rightarrow r $                    & 7,000                & $ c \rightarrow e $                      & 3,342                & $ c \rightarrow s $                      & 906                                 \\
                       & $ r{[}s1{]} c \rightarrow (h,t) $      & 16,551               & $ e{[}s1{]} c   \rightarrow (tr,a) $      & 3,342                & $ s{[}s1{]} c   \rightarrow (at,o) $      & 953                         \\
                       & $ h{[}s1{]}t {[}s2{]}c \rightarrow r $ & 31,736               & $ tr{[}s1{]}a   {[}s2{]}c \rightarrow e $ & 5,426                & $ at{[}s1{]}o   {[}s2{]}c \rightarrow s $ & 1,456                       \\ \cmidrule(lr){2-7}
\multirow{4}{*}{Bidirectional} & $ c \rightarrow (h) $                  & 7,000                & $ c \rightarrow (tr) $                    & 3,342                & $ c \rightarrow (at) $                    & 906                                  \\
                       & $ c \rightarrow (t) $                  & 7,000                & $ c \rightarrow (a) $                    & 3,342                & $ c \rightarrow (o) $                     & 906                                   \\
                       & $ h{[}s1{]}c \rightarrow (t,r) $       & 11,278               & $ tr{[}s1{]}c   \rightarrow (e,a) $       & 3,342                & $ at{[}s1{]}c   \rightarrow (o,s) $       & 1,272                     \\
                       & $ t {[}s1{]}c \rightarrow (h,r) $      & 30,609               & $ at{[}s1{]}c   \rightarrow (tr,e) $      & 5,426                & $ o {[}s1{]}c   \rightarrow (at,s) $      & 1,258               
\\ \bottomrule
\end{tabular}}
\vspace{-0.2cm}
\caption{Training sample statistics after data decomposition on three task datasets. ``$\rightarrow$'' denotes the model's data flow, connecting the model inputs (left of the arrow) to the outputs (right of the arrow). The tokens $[s1]$ and $[s2]$ serve as separators between multiple input contents. Training scale for instruction tuning shown in Appendix~\ref{app:B}.}
\vspace{-0.1cm}
\label{tab:2}
\end{table*}

\begin{table*}[!bt]
\centering
\scalebox{0.74}{
\begin{tabular}{p{4.8cm}p{1.8cm}p{1.1cm}<{\centering}p{1.1cm}<{\centering}p{1.1cm}<{\centering}p{1.1cm}<{\centering}p{1.1cm}<{\centering}p{1.1cm}<{\centering}p{1.1cm}<{\centering}p{1.1cm}<{\centering}p{1.1cm}<{\centering}}
\toprule
\textbf{Objective}                  & \textbf{Planner}         & \textbf{a}     & \textbf{b}     & \textbf{c}    & \textbf{d}    & \textbf{e}    & \textbf{f}    & \textbf{g}     & \textbf{h}    & \textbf{i}    \\ \midrule
\multirow{2}{*}{Performance}      & $ f(x)_{dataset} $ & -1.75  & -3.49  & -2.07 & 2.11 & 0.61  & 0.14 & 1.47  & 39.26  & 26.45  \\
                           & $ f(x)_{model} $   & -6.30  & -10.00 & -7.70  & 6.90  & 8.00 & 10.00  & -9.80 & 0.50 & -0.20  \\ \cmidrule{1-11}
\multirow{2}{*}{Performance + Efficiency} & $ f(x)_{dataset} $ & -1.30   & -1.50 & -3.70 & 1.90 & 4.70  & 0.70  & 23.70  & 1.20 & 6.60  \\
                           & $ f(x)_{model} $   & 7.90 & -3.50  & 8.70 & -2.70 & 9.80 & 8.10  & 29.60  & -17.30  & 7.50 \\ \bottomrule
\end{tabular}}
\vspace{-0.2cm}
\caption{The coefficient of the DTI planner in different objectives, where the variables' mapping range is 1 to 10.} 
\vspace{-0.2cm}
\label{tab:3}
\end{table*}

\section{Experiments}\label{6}

\subsection{Experimental Setup}\label{6.1}

\textbf{Datasets.} 
We conduct the experiments on three public datasets, including CMeIE-V2\footnote{\url{https://tianchi.aliyun.com/dataset/95414}}~\cite{zhang2022cblue},
ACE05~\cite{walker2006ace},
and 14Lap~\cite{xu2020position}. 
The statistics of these datasets are reported in Table~\ref{tab:1}, and the training sample statistics after task decomposition are listed in Table~\ref{tab:2}. 
Besides, since the gold references for the CMeIE-V2 test set are not publicly available, we evaluate models on the validation set.

\textbf{Models and hyperparameters.}
We experiment on open-source LLMs with various parameter sizes, including Qwen3-8B~\cite{qwen3}, InternLM3-8B~\cite{cai2024internlm2}, LLaMA3.1-8B~\cite{dubey2024llama}, and GLM4-9B~\cite{glm2024chatglm}. 
The parameter settings for tuning the LLMs are detailed in Appendix~\ref{app:A}.
We also test closed-source LLMs via API. Based on three trials, the average costs per run are \$0.0036 for GPT-5.1, \$0.0016 for QwQ-MAX, and \$0.00037 for DeepSeek-R1.
The coefficients of the DTI planner are solved, as shown in Table~\ref{tab:3}. 
The analysis of coefficient solution non-uniqueness and the variable mapping range values are provided in Appendix~\ref{app:FF} and Appendix~\ref{app:F}, respectively.

\begin{table}[!bt]
\centering
\scalebox{0.668}{
\begin{tabular}{p{1.4cm}p{2.1cm}p{1.2cm}<{\centering}p{0.8cm}<{\centering}p{0.8cm}<{\centering}p{1.cm}<{\centering}p{1.2cm}<{\centering}}
\toprule
\multicolumn{2}{l}{\textbf{Method}}   & \textbf{Time(s)}\textcolor{red}{$\downarrow$} & \textbf{P(\%)} & \textbf{R(\%)} & \textbf{F1(\%)}\textcolor{red}{$\uparrow$} & \textbf{Sea.(s)}\textcolor{red}{$\downarrow$}\\ \midrule
  \rowcolor{gray!12}\multicolumn{7}{c}{\textbf{\textit{Single-objective}}}       \\
Random               &             & 66,122         & 53.16 & 49.45 & 51.24   & 6               \\
\multicolumn{2}{l}{Grid Search}            & 96,593         & 65.41 & 46.42 & \textbf{54.30}   & 742,090         \\
\multicolumn{2}{l}{Greedy Search}   & 113,262        & 63.58 & 45.36 & 52.94   & 430,599         \\
\multicolumn{2}{l}{Hill-climbing Search}             & 96,593         & 65.41 & 46.42 & 54.30   & 736,082         \\ \hdashline
MetaGPT              & \textit{GPT-5.1}      & 78,212         & 62.32 & 43.96 & 51.55   & 164             \\
                     & \textit{QwQ-MAX}      & 90,352         & 44.79 & 60.82 & 51.59   & 165             \\
                     & \textit{Deepseek-R1}  & 113,262        & 63.58 & 45.36 & 52.94   & 164             \\ \hdashline
\multicolumn{2}{l}{\textbf{PlanE (ours)}}         & 96,593         & 65.41 & 46.42 & \textbf{54.30}   & \textbf{187,257}         \\  \cmidrule{1-7}
  \rowcolor{gray!12}\multicolumn{7}{c}{\textbf{\textit{Multi-objective}}}          \\
Random               &             & 66,122         & 53.16 & 49.45 & 51.24   & 6               \\
\multicolumn{2}{l}{Grid Search}      & \textbf{79,593}         & 63.10 & 46.10 & \textbf{53.28}   & 742,090         \\
\multicolumn{2}{l}{Greedy Search}  & 11,405         & 50.76 & 51.41 & 51.09   & 187,257         \\
\multicolumn{2}{l}{Hill-climbing Search}      & 96,593         & 65.41 & 46.42 & 54.30   & 736,082         \\ \hdashline
MetaGPT              & \textit{GPT-5.1}      & 11,167         & 50.83 & 50.70 & 50.76   & 182             \\
                     & \textit{QwQ-MAX}      & 96,593         & 45.90 & 59.32 & 51.75   & 185             \\
                     & \textit{Deepseek-R1}  & 68,227         & 51.34 & 51.25 & 51.29   & 181             \\ \hdashline
\multicolumn{2}{l}{\textbf{PlanE (ours)}}     & \textbf{79,593}         & 63.10 & 46.10 & \textbf{53.28}   & \textbf{187,257}         \\ \bottomrule
\end{tabular}}
\vspace{-0.2cm}
\caption{Comparison of single-objective (Performance) and multi-objective (Performance + Efficiency) optimization the on Qwen3-8B in CMeIE-V2. ``Time'' = model construction time, ``Sea.'' = search time for the optimal DTI combination.}
\vspace{-0.5cm}
\label{tab:5555}
\end{table}

\begin{table*}[!bt]
\centering
\scalebox{0.7}{
\begin{tabular}{p{0.5cm}<{\centering}p{2.4cm}p{2.2cm}p{2.2cm}<{\centering}p{1.3cm}<{\centering}p{1.1cm}<{\centering}p{1.1cm}<{\centering}p{1.1cm}<{\centering}p{0.4cm}<{\centering}p{0.4cm}<{\centering}p{0.4cm}<{\centering}p{1.6cm}<{\centering}p{1.6cm}<{\centering}}
\toprule
\multirow{2}{*}{\textbf{ID}} & \multicolumn{3}{c}{\textbf{Combination}}                                       & \multirow{2}{*}{\textbf{Time(s)}} & \multirow{2}{*}{\textbf{P(\%)}} & \multirow{2}{*}{\textbf{R(\%)}} & \multirow{2}{*}{\textbf{F1(\%)}} & \multicolumn{3}{c}{\textbf{Variable}} & \multirow{2}{*}{$ f(x)_{dataset} $} & \multirow{2}{*}{$ f(x)_{model} $} \\ \cmidrule(lr){2-4} \cmidrule(lr){9-11}
                    & Data                   & Tuning                     & Inference &                          &                      &                      &                     & $ vd $ & $ vt $ & $ vi $ &                                     &                                   \\ \midrule

                     
  \rowcolor{gray!12}  1                   &                          & SFT                      & Direct                                 & 11,405                                        & 50.76                                     & 51.41                                     & 51.09                                    & 8                                   & 10                                 & 1                                   & 143                                                      & -769                                                   \\
2                                   &                          & SFT + GRPO                   & Direct                                 & 17,342                                        & 49.95                                     & 50.36                                     & 50.15                                    & 8                                   & 1                                  & 1                                   & -18                                                      & -370                                                   \\
3                                   &                          & SFT + DPO                   & Direct                                 & 11,167                                        & 50.83                                     & 50.70                                      & 50.76                                    & 8                                   & 7                                  & 1                                   & 152                                                      & -456                                                   \\
4                                   & \multirow{-4}{*}{None}     & SFT + KTO                   & Direct                                 & 13,931                                        & 51.66                                     & 50.11                                     & 50.87                                    & 8                                   & 8                                  & 1                                   & 156                                                      & -540    \\  \cmidrule(lr){2-13}                                               
5                                   &                          &                         & Direct                                 & 68,227                                        & 51.34                                     & 51.25                                     & 51.29                                    & 10                                  & 10                                 & 1                                   & 126                                                      & -861                                                   \\
6                                   &                          &                         & Intersection                                  & 70,029                                        & 63.30                                      & 45.14                                     & 52.70                                     & 10                                  & 10                                 & 10                                  & 227                                                      & -5                                                     \\
7                                   &                          & \multirow{-3}{*}{SFT}    & Union                                  & 70,029                                        & 45.49                                     & 58.10                                      & 51.03                                    & 10                                  & 10                                 & 2                                   & 154                                                      & -704                                                   \\  \cmidrule(lr){3-13}  
8                                   &                          &                         & Direct                                 & 111,470                                       & 52.34                                     & 51.99                                     & 52.17                                    & 10                                  & 1                                  & 1                                   & -73                                                      & -586                                                   \\
9                                   &                          &                         & Intersection                                  & 113,262                                       & 63.58                                     & 45.36                                     & 52.94                                    & 10                                  & 1                                  & 10                                  & 16                                                       & -540                                                   \\
10                                  &                          & \multirow{-3}{*}{SFT + GRPO} & Union                                  & 113,262                                       & 46.13                                     & 58.11                                     & 51.43                                    & 10                                  & 1                                  & 2                                   & -46                                                      & -520                                                   \\ \cmidrule(lr){3-13}  
11                                  &                          &                         & Direct                                 & 77,641                                        & 52.42                                     & 50.89                                     & 51.64                                    & 10                                  & 7                                  & 1                                   & 123                                                      & -589                                                   \\
12                                  &                          &                         & Intersection                                  & 79,593                                        & 63.10                                      & 46.10                                      & 53.28                                    & 10                                  & 7                                  & 10                                  & 220                                                      & -4                                                     \\
13                                  &                          & \multirow{-3}{*}{SFT + DPO} & Union                                  & 79,593                                        & 44.78                                     & 59.83                                     & 51.22                                    & 10                                  & 7                                  & 2                                   & 150                                                      & -463                                                   \\ \cmidrule(lr){3-13}  
14                                  &                          &                         & Direct                                 & 94,879                                        & 51.78                                     & 51.63                                     & 51.70                                     & 10                                  & 8                                  & 1                                   & 131                                                      & -660                                                   \\
\rowcolor{lightgreen!85} 15 &                          &                         & Intersection & 96,593       & \textbf{65.41}    & {46.42}    & \textbf{54.30}    & 10 & {8} & {10} & \textbf{229}                     & \textbf{16}                    \\
16                                  & \multirow{-13}{*}{Pipeline-based} & \multirow{-3}{*}{SFT + KTO} & Union                                  & 96,593                                        & 45.90                                      & 59.32                                     & 51.75                                    & 10                                  & 8                                  & 2                                   & 158                                                      & -523                                                   \\  \cmidrule(lr){2-13}
17                                  &                          &                         & Direct                                 & 65,185                                        & 50.39                                     & 50.33                                     & 50.36                                    & 1                                   & 10                                 & 1                                   & 91                                                       & -842                                                   \\
18                                  &                          &                         & Intersection                                  & 66,804                                        & 60.61                                     & 46.05                                     & 52.34                                    & 1                                   & 10                                 & 10                                  & 142                                                      & -634                                                   \\
19                                  &                          & \multirow{-3}{*}{SFT}    & Union                                  & 66,804                                        & 45.59                                     & 58.58                                     & 51.27                                    & 1                                   & 10                                 & 2                                   & 113                                                      & -757                                                   \\ \cmidrule(lr){3-13}  
20                                  &                          &                         & Direct                                 & 110,836                                       & 51.11                                     & 50.48                                     & 50.79                                    & 1                                   & 1                                  & 1                                   & 63                                                       & -9                                                     \\
21                                  &                          &                         & Intersection                                  & 112,490                                       & 60.95                                     & 45.57                                     & 52.15                                    & 1                                   & 1                                  & 10                                  & 103                                                      & -611                                                   \\
22                                  &                          & \multirow{-3}{*}{SFT + GRPO} & Union                                  & 112,490                                       & 45.45                                     & 58.38                                     & 51.11                                    & 1                                   & 1                                  & 2                                   & 84                                                       & -14                                                    \\ \cmidrule(lr){3-13}  
23                                  &                          &                         & Direct                                 & 76,539                                        & 50.72                                     & 50.22                                     & 50.47                                    & 1                                   & 7                                  & 1                                   & 144                                                      & -384                                                   \\
24                                  &                          &                         & Intersection                                  & 78,212                                        & 62.32                                     & 43.96                                     & 51.55                                    & 1                                   & 7                                  & 10                                  & 192                                                      & -446                                                   \\
25                                  &                          & \multirow{-3}{*}{SFT + DPO} & Union                                  & 78,212                                        & 43.52                                     & 58.81                                     & 50.02                                    & 1                                   & 7                                  & 2                                   & 166                                                      & -330                                                   \\ \cmidrule(lr){3-13}  
26                                  &                          &                         & Direct                                 & 88,759                                        & 52.58                                     & 50.75                                     & 51.65                                    & 1                                   & 8                                  & 1                                   & 133                                                      & -517                                                   \\
27                                  &                          &                         & Intersection                                  & 90,352                                        & 63.34                                     & 44.90                                      & 52.55                                    & 1                                   & 8                                  & 10                                  & 182                                                      & -489                                                   \\
28                                  & \multirow{-13}{*}{Bidirectional} & \multirow{-3}{*}{SFT + KTO} & Union                                  & 90,352                                        & 44.79                                     & 60.82                                     & 51.59                                    & 1                                   & 8                                  & 2                                   & 155                                                      & -452                      \\ \bottomrule
\end{tabular}}
\vspace{-0.2cm}
\caption{The prediction results of constructing the DTI planner on the Qwen3-8B in CMeIE-V2.}
\vspace{-0.4cm}
\label{tab:4}
\end{table*}

\begin{table*}[t]
\centering
\scalebox{0.7}{
\begin{tabular}{p{0.5cm}<{\centering}p{2.4cm}p{2.2cm}p{2.2cm}<{\centering}p{1.3cm}<{\centering}p{1.1cm}<{\centering}p{1.1cm}<{\centering}p{1.1cm}<{\centering}p{0.4cm}<{\centering}p{0.4cm}<{\centering}p{0.4cm}<{\centering}p{1.6cm}<{\centering}p{1.6cm}<{\centering}}
\toprule
\multirow{2}{*}{\textbf{ID}} & \multicolumn{3}{c}{\textbf{Combination}}                                       & \multirow{2}{*}{\textbf{Time(s)}} & \multirow{2}{*}{\textbf{P(\%)}} & \multirow{2}{*}{\textbf{R(\%)}} & \multirow{2}{*}{\textbf{F1(\%)}} & \multicolumn{3}{c}{\textbf{Variable}} & \multirow{2}{*}{$ f(x)_{dataset} $} & \multirow{2}{*}{$ f(x)_{model} $} \\ \cmidrule(lr){2-4} \cmidrule(lr){9-11}
                    & Data                   & Tuning                    & Inference &                          &                      &                      &                     & $ vd $ & $ vt $ & $ vi $ &                                     &                                   \\ \midrule
  \rowcolor{gray!12}   1                          &                          & SFT                      & Direct                         & 11,405                                        & 50.76                                     & 51.41                                     & 51.09                                    & 8                          & 10                         & 1                          & 167.9                                                    & 179.0                                                    \\
2                          &                          & SFT + GRPO                   & Direct                         & 17,342                                        & 49.95                                     & 50.36                                     & 50.15                                    & 8                          & 1                          & 1                          & 162.5                                                    & 802.7                                                  \\
3                          &                          & SFT + DPO                   & Direct                         & 11,167                                        & 50.83                                     & 50.70                                      & 50.76                                    & 8                          & 7                          & 1                          & 193.1                                                    & 449.9                                                  \\
4                          & \multirow{-4}{*}{None}     & SFT + KTO                   & Direct                         & 13,931                                        & 51.66                                     & 50.11                                     & 50.87                                    & 8                          & 8                          & 1                          & 187.7                                                    & 366.6      \\  \cmidrule(lr){2-13}                                            
5                          &                          &                         & Direct                         & 68,227                                        & 51.34                                     & 51.25                                     & 51.29                                    & 10                         & 10                         & 1                          & 215.9                                                    & 488.2                                                  \\
\rowcolor{lightyellow!40} 6  &                          &                         & Intersection & 70,029                & 63.30              & 45.14             & \textbf{52.70}\textcolor{red}{$^4$}             & 10 & 10 & 10 & \textbf{395.0}                     & 3,028.0                       \\
7                          &                          & \multirow{-3}{*}{SFT}    & Union                         & 70,029                                        & 45.49                                     & 58.10                                      & 51.03                                    & 10                         & 10                         & 2                          & 265.4                                                    & 700.8                                                  \\ \cmidrule(lr){3-13} 
8                          &                          &                         & Direct                         & 111,470                                       & 52.34                                     & 51.99                                     & 52.17                                    & 10                         & 1                          & 1                          & 176.3                                                    & 1,160.5                                               \\
9                          &                          &                         & Intersection                         & 113,262                                       & 63.58                                     & 45.36                                     & \textbf{52.94}\textcolor{red}{$^3$}                                   & 10                         & 1                          & 10                         & 298.7                                                    & 3,044.2                                               \\
10                         &                          & \multirow{-3}{*}{SFT + GRPO} & Union                         & 113,262                                       & 46.13                                     & 58.11                                     & 51.43                                    & 10                         & 1                          & 2                          & 219.5                                                    & 1,300.2                                               \\ \cmidrule(lr){3-13} 
11                         &                          &                         & Direct                         & 77,641                                        & 52.42                                     & 50.89                                     & 51.64                                    & 10                         & 7                          & 1                          & 229.7                                                    & 775.3                                                  \\
\rowcolor{lightblue!15} 12 &                          &                         & Intersection & 79,593                &  63.10              & 46.10              & \textbf{53.28}\textcolor{red}{$^2$}           & 10 & 7  & 10 & 389.9                            & \textbf{3,096.4}              \\
13                         &                          & \multirow{-3}{*}{SFT + DPO} & Union                         & 79,593                                        & 44.78                                     & 59.83                                     & 51.22                                    & 10                         & 7                          & 2                          & 277.1                                                    & 963.6                                                  \\ \cmidrule(lr){3-13} 
14                         &                          &                         & Direct                         & 94,879                                        & 51.78                                     & 51.63                                     & 51.70                                     & 10                         & 8                          & 1                          & 228.1                                                    & 686.6                                                  \\
15                         &                          &                         & Intersection                         & 96,593                                        & 65.41                                     & 46.42                                     & \textbf{54.30}\textcolor{red}{$^1$}                                     & 10                         & 8                          & 10                         & 394.6                                                    & 3,080.6                                               \\
16                         & \multirow{-13}{*}{Pipeline-based} & \multirow{-3}{*}{SFT + KTO} & Union                         & 96,593                                        & 45.90                                      & 59.32                                     & 51.75                                    & 10                         & 8                          & 2                          & 276.2                                                    & 883.0                                                   \\  \cmidrule(lr){2-13}
17                         &                          &                         & Direct                         & 65,185                                        & 50.39                                     & 50.33                                     & 50.36                                    & 1                          & 10                         & 1                          & -82.0                                                      & -405.5                                                 \\
18                         &                          &                         & Intersection                         & 66,804                                        & 60.61                                     & 46.05                                     & 52.34                                    & 1                          & 10                         & 10                         & -283.6                                                   & 1,340.5                                               \\
19                         &                          & \multirow{-3}{*}{SFT}    & Union                         & 66,804                                        & 45.59                                     & 58.58                                     & 51.27                                    & 1                          & 10                         & 2                          & -74.8                                                    & -281.1                                                 \\ \cmidrule(lr){3-13} 
20                         &                          &                         & Direct                         & 110,836                                       & 51.11                                     & 50.48                                     & 50.79                                    & 1                          & 1                          & 1                          & 32.3                                                     & 48.1                                                   \\
21                         &                          &                         & Intersection                         & 112,490                                       & 60.95                                     & 45.57                                     & 52.15                                    & 1                          & 1                          & 10                         & -226.0                                                     & 1,138.0                                              \\
22                         &                          & \multirow{-3}{*}{SFT + GRPO} & Union                         & 112,490                                       & 45.45                                     & 58.38                                     & 51.11                                    & 1                          & 1                          & 2                          & 33.2                                                     & 99.6                                                   \\ \cmidrule(lr){3-13} 
23                         &                          &                         & Direct                         & 76,539                                        & 50.72                                     & 50.22                                     & 50.47                                    & 1                          & 7                          & 1                          & -16.9                                                    & -191.3                                                 \\
24                         &                          &                         & Intersection                         & 78,212                                        & 62.32                                     & 43.96                                     & 51.55                                    & 1                          & 7                          & 10                         & -237.4                                                   & 1,336.0                                               \\
25                         &                          & \multirow{-3}{*}{SFT + DPO} & Union                         & 78,212                                        & 43.52                                     & 58.81                                     & 50.02                                    & 1                          & 7                          & 2                          & -11.8                                                    & -91.2                                                  \\ \cmidrule(lr){3-13} 
26                         &                          &                         & Direct                         & 88,759                                        & 52.58                                     & 50.75                                     & 51.65                                    & 1                          & 8                          & 1                          & -35.6                                                    & -255.7                                                 \\
27                         &                          &                         & Intersection                         & 90,352                                        & 63.34                                     & 44.90                                      & 52.55                                    & 1                          & 8                          & 10                         & -249.8                                                   & 1,344.5                                               \\
28                         & \multirow{-13}{*}{Bidirectional} & \multirow{-3}{*}{SFT + KTO} & Union                         & 90,352                                        & 44.79                                     & 60.82                                     & 51.59                                    & 1                          & 8                          & 2                          & -29.8                                                    & -147.5                                                
\\ \bottomrule
\end{tabular}}
\vspace{-0.2cm}
\caption{The prediction results of constructing the DTI planner on multi-objective optimization. Since Performance (F1) and Efficiency (training time, $T$) operate on different scales, we apply min-max normalization to both metrics.
The final score $S$ for a DTI combination is the weighted sum of the normalized components:
$S = \alpha \cdot \text{Norm}(F_1) + \beta \cdot \text{Norm}(T)$ (In our default setup, $\alpha = \beta = 0.5$). This joint score $S$ replaces the single-objective in the ranking stage. All subsequent steps—including the search and the planner's loss calculation—remain identical to the single-objective pipeline described. The red numerical labels indicate the best performance ranking.} 
\label{tab:5}
\end{table*}

\begin{table*}[!bt]
\centering
\scalebox{0.7}{
\begin{tabular}{p{1.9cm}p{2.6cm}p{1.0cm}<{\centering}p{1.0cm}<{\centering}p{1.0cm}<{\centering}p{1.2cm}<{\centering}p{1.2cm}<{\centering}p{1.2cm}<{\centering}p{1.0cm}<{\centering}p{1.0cm}<{\centering}p{1.0cm}<{\centering}p{1.0cm}<{\centering}p{1.2cm}<{\centering}}
\toprule
\multirow{2}{*}{\textbf{Dataset}} & \multirow{2}{*}{\textbf{Model}} & \multicolumn{3}{c}{\textbf{Data}}       & \multicolumn{4}{c}{\textbf{Tuning}} & \multicolumn{3}{c}{\textbf{Inference}} & \multirow{2}{*}{\textbf{Best}} \\  \cmidrule(l){3-5} \cmidrule(l){6-9} \cmidrule(l){10-12}
                    &    & None    & Pipe. & Bidi. & SFT          & S.+ G.   & S.+ D. & S.+ K.     & Direct & Int. & Union &                      \\ \midrule
\multirow{4}{*}{CMeIE-V2} & Qwen3-8B               & 51.09  & 51.29  & 51.75  & 51.09 & 50.15 & 50.76 & 50.87 & 50.83  & 52.52        & 51.15 & \textbf{54.30}                 \\
& GLM4-9B                & 49.99  & 53.07  & 50.90   & 49.99 & 50.59 & 49.15 & 50.35 & 52.11  & 53.32        & 52.11 & \textbf{54.03}               \\
& LLama3.1-8B            & 51.06  & 50.87  & 51.46  & 51.06 & 50.56 & 49.45 & 48.70  & 51.02  & 52.11        & 50.93 & \textbf{52.62}                \\
& InternLM3-8B           & 48.12  & 49.87  & 49.59  & 48.12 & 47.42 & 47.50  & 47.22 & 49.51  & 50.45        & 50.25 & \textbf{50.77}                           \\  \cmidrule(l){1-13}
\multirow{4}{*}{ACE05} & Qwen3-8B               & 46.06  & 51.98  & 52.57  & 46.06 & 45.74 & 45.46 & 45.19 & 51.60   & 52.96        & 52.90  & \textbf{54.88}                \\
& GLM4-9B                & 50.52  & 53.41  & 52.23  & 50.52 & 52.02 & 52.86 & 52.13 & 51.71  & 52.59        & 52.45 & \textbf{54.77}                \\
& LLama3.1-8B            & 48.51  & 49.42  & 50.17  & 48.51 & 48.95 & 50.19 & 49.47 & 49.95  & 50.90         & 50.71 & \textbf{53.27}                \\
& InternLM3-8B           & 42.74  & 49.52  & 50.13  & 42.74 & 42.02 & 42.19 & 44.87 & 48.25  & 51.18        & 49.68 & \textbf{52.98}                \\  \cmidrule(l){1-13}
\multirow{4}{*}{14Lap} & Qwen3-8B               & 60.68  & 61.52  & 61.98  & 60.68 & 61.58 & 59.58 & 61.22 & 61.58  & 63.15        & 62.17 & \textbf{65.96}                \\
& GLM4-9B                & 62.24  & 63.09  & 63.24  & 62.24 & 65.22 & 61.23 & 62.13 & 63.14  & 64.54        & 61.81 & \textbf{68.32}                \\
& LLama3.1-8B            & 62.00     & 61.80   & 62.72  & 62.00    & 62.06 & 60.51 & 59.35 & 61.82  & 63.38        & 62.22 & \textbf{64.26}                \\
& InternLM3-8B           & 56.56  & 59.87  & 60.30   & 56.56 & 58.17 & 57.78 & 57.33 & 59.48  & 62.46        & 61.23 & \textbf{64.16}              \\ \bottomrule
\end{tabular}}
\vspace{-0.2cm}
\caption{Ablation study (F1\%). ``Pipe.'', ``Bidi.'', ``S.'', ``G.'', ``D.'', ``K.'', ``Int.'', and  ``Best'' denote the ``Pipeline-based'', ``Bidirectional'', ``SFT'', ``GRPO'', ``DPO'', ``KTO '', ``Intersection'' and ``the best performance'', respectively.} 
\vspace{-0.3cm}
\label{tab:6}
\end{table*}

\textbf{Evaluation metrics.}
Following the previous works~\cite{zhang2024bidirectional}, we employ three standard evaluation metrics, i.e., Precision (P), Recall (R), and strict Micro-F1 (F1). Notably, in RE, a triplet is considered correct only if the relation type, along with the types and the boundaries of the head-tail entities are precisely determined. For EE, we decompose events into (trigger, event type, argument, argument role) quadruples, requiring exact matching of all components. For ABSA, a triplet is regarded as correct only when the aspect, opinion, and sentiment all coincide with the ground truth.


\subsection{Exp-I: Main results}\label{6.2}
For a fair comparison, we build an LLM-Planner by prompting MetaGPT with PlanE's empirical data and redefining its operating procedure to select DTI combinations. 
As listed in Table \ref{tab:5555}, we choose three base-LLMs (GPT-5.1, QwQ-MAX, DeepSeek-R1) as the LLM-Planner.
We observe that: 1) PlanE consistently outperforms all baselines under both single- and multi-objective optimization. 
2) In both single-objective and multi-objective optimization settings, PlanE achieves performance comparable to Grid Search, while reducing search time by 554,833 seconds.






\subsection{Exp-II: How is the generalization ability of the DTI planner?}\label{6.3}
\textit{\textbf{View 1: Same task dataset, different base-LLMs.}}
To verify the generalization of the proposed PlanE, we design the View 1 experiment: fitting the DTI function on several base-LLMs using the same task dataset. The planner $ f(x)_{dataset} $ is fitted on CMeIE-V2 with three base-LLMs: InternLM3-8B, LLaMA-3.1-8B, and GLM4-9B (detailed fitted data are shown in Appendix~\ref{app:D}). The results are reported in Table \ref{tab:3}. To further assess the planner’s validity, we evaluate it on Qwen3-8B, with the results shown in Table \ref{tab:4}. The DTI planner identifies the following best DTI combination: pipeline-based data decomposition, SFT + KTO tuning, and intersection inference. The resulting extractive-based LLM achieves the best performance among the 28 candidate DTI combinations under single-objective optimization focused solely on performance.





\noindent\textit{\textbf{View 2: Same base-LLM, different task datasets.}}
To evaluate the effectiveness of PlanE from the View 2 experiment, we conduct an experiment in which the DTI planer is fitted on multiple task datasets using the same base-LLM. The planner $f(x)_{model}$ is fitted using the results from EE and an ABSA task dataset based on Qwen3-8B (detailed fitted data are shown in Appendix~\ref{app:E}), and the corresponding coefficient results are reported in Table \ref{tab:3}. To verify the planner's generalization, we further test it on the RE dataset CMeIE-V2, with the results listed in Table \ref{tab:4}. The planner identifies the following best DTI combination: pipeline-based data decomposition, SFT + KTO tuning, and intersection inference. The resulting extractive-based LLM (ID=15) outperforms the other 27 candidate DTI combinations under single-objective optimization focused on performance.

\subsection{Exp-III: Can the DTI planner be used for multi-objective optimization?}\label{6.4}
To evaluate the generalization capability of the DTI planner, we further apply it to multi-objective optimization. When constructing extractive-based LLMs, we sometimes need to balance both performance and efficiency, measuring efficiency by the combined training and inference time. The DTI planner is still constructed from two views, and its coefficients are listed in Table \ref{tab:3}. The prediction results of the planner are shown in Table \ref{tab:5}. 
Experimental results show that the DTI planner effectively achieves multi-objective optimization.
Specifically, from the view of the same dataset $f(x)_{dataset}$, the optimal choice for multi-objective optimization is ID=6 (highlighted in yellow in Table \ref{tab:5}); from the view of the same base-LLM $f(x)_{model}$, the optimal choice is ID=12 (highlighted in blue in Table \ref{tab:5}).

\subsection{Exp-IV: Does DTI impact the performance of extractive-based LLMs?}\label{6.5}
In this subsection, we conduct ablation experiments to examine how the three impact elements: Data, Tuning, and Inference, each contribute to building an extractive-based LLM. The ablation study spans three tasks, including: RE task (e.g., CMeIE-V2), EE task (e.g., ACE05), and ABSA task (e.g., 14Lap), and across four base-LLMs. The results, summarized in Table \ref{tab:6}, reveal that each factor affects various base-LLMs to varying extents, and their effectiveness across datasets shows no pattern.

\section{Conclusion}
\label{Conclusion and Limitations}
In this paper, we propose PlanE, a framework that is the first to focus on extractive-based LLM construction. We design a DTI planner to select the optimal DTI combination for specific datasets/base-LLMs to improve construction efficiency. We conducted experimental analyses of PlanE’s generalizability from two views: (1) across different datasets using the same base-LLM, and (2) on the same dataset using different base-LLMs.  
Moreover, results show that PlanE improves the F1-score over MetaGPT by 1.36\% (single-objective), and matches the performance of Grid Search under single- and multi-objective optimization while saving 554,833 seconds of search time.
\section*{Limitations}
While PlanE offers a valuable solution for planning of Data, Tuning, and Inference for extractive-based LLMs, it has three main limitations: 

1) The DTI planner is a discrete function and cannot be directly optimized using gradient information, which can be addressed through reinforcement learning methods~\cite{kaelbling1996reinforcement}.

2) We do not consider all influencing factors, particularly practical constraints such as computational resources, or additional DTI combinations involving more complex and diverse data, tuning, and inference. Therefore, we leave this exploration for future work. 


3) We currently employ scalar encoding for the DTI planner. Although it carries less information than vectorized alternatives, empirical results show it provides sufficient structural cues to stably reach the optimal performance (ID 15 in Table~\ref{tab:4}). Thus, the simplified representation does not bottleneck the current model's efficacy. Nevertheless, exploring vectorized representations remains a key future work to improve search efficiency and minimize the overhead of computing initial DTI combinations.

\section*{Acknowledgments}
This work is supported by the Shanghai Natural Science Foundation Project under Grant 25ZR1402116.


\bibliography{custom}

\newpage
\appendix


\renewcommand{\thetable}{A\arabic{table}}
\renewcommand{\thefigure}{A\arabic{figure}}
\setcounter{figure}{0}
\setcounter{table}{0}

\section{Parameter Settings}\label{app:A}
For model fine-tuning, we utilize Low-Rank Adaptation (LoRA) with the following configuration: a batch size of 32, a learning rate of \(1 \times 10^{-4}\), and 4 epochs. To optimize computational efficiency, we employ the bfloat16 data type. The LoRA-specific hyperparameters are set as follows: the rank of the low-rank matrices (\(lora\_r\)) is 16, and the scaling factor (\(lora\_\alpha\)) is 32.

For model reinforcement learning, we use the GRPO, DPO, and KTO methods and utilize Low-Rank Adaptation (LoRA) with the following configuration: a batch size of 8, a generation number of 8,  a learning rate of \(1 \times 10^{-6}\), and 2 epochs. To optimize computational efficiency, we employ the bfloat16 data type. The LoRA-specific hyperparameters are set as follows: the rank of the low-rank matrices (\(lora\_r\)) is 16, and the scaling factor (\(lora\_\alpha\)) is 32.

Our experiments are conducted on a workstation running Ubuntu 20.04.6 LTS, with two Intel(R) Xeon(R) Platinum 8336C CPUs, four NVIDIA A800 GPUs, and 1.0TiB of memory.


During inference, we configure the \(temperature\) to 0.3 and the top-p sampling parameter (\(top\_p\)) to 0.9, ensuring a balance between diversity and coherence in the generated outputs.

\section{Training Scale for Different Instruction Tunings}\label{app:B}
This section presents the number of training samples for different instruction tuning strategies across all three datasets. We provide a detailed breakdown of the sample sizes for each decomposition approach, offering insight into the distribution of data used for training under various configurations, as shown in Table~\ref{tab:app_B}. 


\begin{table*}[t]
\centering
\scalebox{0.72}{
\begin{tabular}{p{2.5cm}p{2.5cm}p{12.9cm}p{2.5cm}<{\centering}}
\toprule
\multicolumn{2}{c}{\textbf{Combination}}               & \multirow{2}{*}{\textbf{Data Structure}}                                                                                                                                  &   \multirow{2}{*}{\textbf{Train Samples}}  \\ \cmidrule{1-2}
Data                   & Tuning                   &                                                                                                                                                             &                         \\ \midrule
  \rowcolor{gray!10}\multicolumn{4}{c}{\textbf{\textit{CMeIE-V2}}}                                                                                                                                                                                               \\
\multirow{3}{*}{None}    & SFT                    & Step1 SFT: $ c \rightarrow (h,r,t) $;                                                                                                                             & 7,000                   \\
                       & \multirow{2}{*}{SFT + RL} & Step1 SFT: $ c \rightarrow (h,r,t) $;                                                                                                                             & 6,000                   \\ 
                       &                        & Step2 RL: $ c \rightarrow (h,r,t) $;                                                                                                                            & 1,000                   \\ \hdashline
\multirow{3}{*}{Pipeline-based} & SFT                     & Step1 SFT: $ c \rightarrow (h,r,t) $; $ c \rightarrow (h,t) $; $ c \rightarrow r $;  $ r{[}s1{]} c \rightarrow (h,t) $;  $ h{[}s1{]}t {[}s2{]}c \rightarrow r $;  & 69,287                  \\
                       & \multirow{2}{*}{SFT + RL} & Step1 SFT: $ c \rightarrow (h,r,t) $; $ c \rightarrow (h,t) $; $ c \rightarrow r $;  $ r{[}s1{]} c \rightarrow (h,t) $;  $ h{[}s1{]}t {[}s2{]}c \rightarrow r $;  & 59,287                  \\
                       &                        & Step2 RL: $ c \rightarrow (h,r,t) $; $ c \rightarrow (h,t) $; $ c \rightarrow r $;  $ r{[}s1{]} c \rightarrow (h,t) $;  $ h{[}s1{]}t {[}s2{]}c \rightarrow r $; & 10,000                  \\ \hdashline
\multirow{3}{*}{Bidirectional} & SFT                    & Step1 SFT: $ c \rightarrow (h,r,t) $; $ c \rightarrow (h) $;  $ c \rightarrow (t) $;  $ h{[}s1{]}c \rightarrow (t,r) $; $ t {[}s1{]}c \rightarrow (h,r) $;        & 62,887                  \\
                       & \multirow{2}{*}{SFT + RL} & Step1 SFT: $ c \rightarrow (h,r,t) $; $ c \rightarrow (h) $;  $ c \rightarrow (t) $;  $ h{[}s1{]}c \rightarrow (t,r) $; $ t {[}s1{]}c \rightarrow (h,r) $;        & 52,887                  \\
                       &                        & Step2 RL: $ c \rightarrow (h,r,t) $; $ c \rightarrow (h) $;  $ c \rightarrow (t) $;  $ h{[}s1{]}c \rightarrow (t,r) $; $ t {[}s1{]}c \rightarrow (h,r) $;       & 10,000                  \\ \cmidrule{1-4}

  \rowcolor{gray!10}\multicolumn{4}{c}{\textbf{\textit{ACE05}}}                                                                                                                                                                                                     \\
\multirow{3}{*}{None}    & SFT                     & Step1 SFT: $ c \rightarrow (tr,e,a) $;                                                                                                                             & 3,342                   \\
                       & \multirow{2}{*}{SFT + RL} & Step1 SFT: $ c \rightarrow (tr,e,a) $;                                                                                                                             & 2,842                   \\
                       &                        & Step2 RL: $ c \rightarrow (tr,e,a) $;                                                                                                                            & 500                     \\ \hdashline
\multirow{3}{*}{Pipeline-based} & SFT                     & Step1 SFT: $ c \rightarrow (tr,e,a) $; $ c \rightarrow (tr,a) $; $ c \rightarrow e $;  $ e{[}s1{]} c \rightarrow (tr,a) $;  $ tr{[}s1{]}a {[}s2{]}c \rightarrow e $;  & 18,794                  \\
                       & \multirow{2}{*}{SFT + RL} & Step1 SFT: $ c \rightarrow (tr,e,a) $; $ c \rightarrow (tr,a) $; $ c \rightarrow e $;  $ e{[}s1{]} c \rightarrow (tr,a) $;  $ tr{[}s1{]}a {[}s2{]}c \rightarrow e $;  & 16,094                  \\
                       &                        & Step2 RL: $ c \rightarrow (tr,e,a) $; $ c \rightarrow (tr,a) $; $ c \rightarrow e $;  $ e{[}s1{]} c \rightarrow (tr,a) $;  $ tr{[}s1{]}a {[}s2{]}c \rightarrow e $; & 2,700                   \\ \hdashline
\multirow{3}{*}{Bidirectional} & SFT                     & Step1 SFT: $ c \rightarrow (tr,e,a) $; $ c \rightarrow (tr) $;  $ c \rightarrow (a) $;  $ tr{[}s1{]}c \rightarrow (e,a) $; $ a {[}s1{]}c \rightarrow (tr,e) $;        & 18,794                  \\
                       & \multirow{2}{*}{SFT + RL} & Step1 SFT: $ c \rightarrow (tr,e,a) $; $ c \rightarrow (tr) $;  $ c \rightarrow (a) $;  $ tr{[}s1{]}c \rightarrow (e,a) $; $ a {[}s1{]}c \rightarrow (tr,e) $;        & 16,094                  \\
                       &                        & Step2 RL: $ c \rightarrow (tr,e,a) $; $ c \rightarrow (tr) $;  $ c \rightarrow (a) $;  $ tr{[}s1{]}c \rightarrow (e,a) $; $ a {[}s1{]}c \rightarrow (tr,e) $;       & 2,700                   \\ \cmidrule{1-4}
  \rowcolor{gray!10}\multicolumn{4}{c}{\textbf{\textit{14lap}}}                                                                                                                                                                                            \\  
  \multirow{3}{*}{None}    & SFT                     & Step1 SFT: $ c \rightarrow (at,o,s) $;                                                                                                                             & 906                   \\
                       & \multirow{2}{*}{SFT + RL} & Step1 SFT: $ c \rightarrow (at,o,s) $;                                                                                                                             & 756                   \\
                       &                        & Step2 RL: $ c \rightarrow (at,o,s) $;                                                                                                                            & 150                     \\ \hdashline
\multirow{3}{*}{Pipeline-based} & SFT                     & Step1 SFT: $ c \rightarrow (at,o,s) $; $ c \rightarrow (at,o) $; $ c \rightarrow s $;  $ s{[}s1{]} c \rightarrow (at,o) $;  $ at{[}s1{]}o {[}s2{]}c \rightarrow s $;  & 5,127                  \\
                       & \multirow{2}{*}{SFT + RL} & Step1 SFT: $ c \rightarrow (at,o,s) $; $ c \rightarrow (at,o) $; $ c \rightarrow s $;  $ s{[}s1{]} c \rightarrow (at,o) $;  $ at{[}s1{]}o {[}s2{]}c \rightarrow s $;  & 4,327                  \\
                       &                        & Step2 RL: $ c \rightarrow (at,o,s) $; $ c \rightarrow (at,o) $; $ c \rightarrow s $;  $ s{[}s1{]} c \rightarrow (at,o) $;  $ at{[}s1{]}o {[}s2{]}c \rightarrow s $;  & 800                   \\ \hdashline
\multirow{3}{*}{Bidirectional} & SFT                     & Step1 SFT: $ c \rightarrow (at,o,s) $; $ c \rightarrow (at) $;  $ c \rightarrow (o) $;  $ at{[}s1{]}c \rightarrow (o,s) $; $ o {[}s1{]}c \rightarrow (at,s) $;        & 5,248                  \\
                       & \multirow{2}{*}{SFT + RL} & Step1 SFT: $ c \rightarrow (at,o,s) $; $ c \rightarrow (at) $;  $ c \rightarrow (o) $;  $ at{[}s1{]}c \rightarrow (o,s) $; $ o {[}s1{]}c \rightarrow (at,s) $;        & 4,448                  \\
                       &                        & Step2 RL: $ c \rightarrow (at,o,s) $; $ c \rightarrow (at) $;  $ c \rightarrow (o) $;  $ at{[}s1{]}c \rightarrow (o,s) $; $ o {[}s1{]}c \rightarrow (at,s) $;        & 800                   
\\ \bottomrule
\end{tabular}}
\caption{Training samples for different data decomposition and tuning strategies for each dataset.}
\label{tab:app_B}
\end{table*}

\begin{table*}[!bt]
\centering
\scalebox{0.72}{
\begin{tabular}{p{1.5cm}<{\centering}p{4.cm}p{14.8cm}}
\toprule
\textbf{Variable}     & \textbf{Combination Element} & \textbf{Value Assignment Rule}                                                                     \\ \midrule
\multirow{3}{*}{\textbf{$vd$}} & None                         & F1 {[}None, SFT, Direct{]}                                                                            \\
                      & Pipeline-based               & F1 {[}Pipeline-based, SFT, Direct{]}                                                                  \\
                      & Bidirectional                & F1 {[}Bidirectional, SFT, Direct{]}                                                                   \\ \cmidrule(l){1-3} 
\multirow{4}{*}{\textbf{$vt$}} & SFT                          & F1 {[}None, SFT, Direct{]}                                                                            \\
                      & SFT + GRPO                   & F1 {[}None, SFT + GRPO, Direct{]}                                                                       \\
                      & SFT + DPO                    & F1 {[}None, SFT + DPO, Direct{]}                                                                        \\
                      & SFT + KTO                    & F1 {[}None, SFT + KTO, Direct{]}                                                                        \\ \cmidrule(l){1-3} 
\multirow{3}{*}{\textbf{$vi$}} & Direct                       & ( F1 {[}None, SFT, Direct{]} + F1 {[}Pipeline-based, SFT, Direct{]} + F1 {[}Bidirectional, SFT, Direct{]} ) / 3 \\
                      & Intersection                 & ( F1 {[}Pipeline-based, SFT, Intersection{]} + F1 {[}Bidirectional, SFT, Intersection{]} ) / 2               \\
                      & Union                        & ( F1 {[}Pipeline-based, SFT, Union{]} + F1 {[}Bidirectional, SFT, Union{]} ) / 2                             \\ \bottomrule
\end{tabular}
}
\caption{The specific principles for mapping DTI strategies to numerical variables. ``F1{[}{]}'' represents the F1 score corresponding to the specific DTI combination.}
\label{tab:app_AA2}
\end{table*}

\begin{table*}[!bt]
\centering
\scalebox{0.67}{
\begin{tabular}{p{1.5cm}<{\centering}p{4.8cm}p{2.5cm}<{\centering}p{2.2cm}<{\centering}p{1.8cm}<{\centering}p{1.0cm}<{\centering}p{2,0cm}<{\centering}p{2.0cm}<{\centering}p{2.0cm}<{\centering}}
\toprule
\multirow{2}{*}{\textbf{Variable}}   & \multirow{2}{*}{\textbf{Strategy}}         & \multicolumn{3}{c}{\textbf{Combination}}   & \multirow{2}{*}{\textbf{F1(\%)}} & \multirow{2}{*}{\textbf{Avg. F1(\%)}} & \multicolumn{2}{c}{\textbf{Mapping Interval}}           \\ \cmidrule(lr){3-5} \cmidrule(l){8-9} 
                        &                                   & Data           & Tuning     & Inference    &                     &                          & $1 \leq v \leq  10 $ & $ 1 \leq v \leq 20 $ \\ \midrule
\multirow{3}{*}{$ vd $} & without Decomposition             & None           & SFT        & Direct       & 51.09               & 51.09                    & 8                          & 16                         \\
                        & Pipeline-based Decomposition      & Pipeline-based & SFT        & Direct       & 51.29               & 51.29                    & 10                         & 20                         \\
                        & Bidirectional Decomposition       & Bidirectional  & SFT        & Direct       & 50.36               & 50.36                    & 1                          & 1                          \\ \cmidrule(l){1-9} 
\multirow{4}{*}{$ vt $} & SFT                               & None           & SFT        & Direct       & 51.09               & 51.09                    & 10                         & 20                         \\
                        & SFT + GRPO                          & None           & SFT + GRPO & Direct       & 50.15               & 50.15                    & 1                          & 1                          \\
                        & SFT + DPO                           & None           & SFT + DPO  & Direct       & 50.76               & 50.76                    & 7                          & 13                         \\
                        & SFT + KTO                           & None           & SFT + KTO  & Direct       & 50.87               & 50.87                    & 8                          & 16                         \\ \cmidrule(l){1-9} 
\multirow{7}{*}{$ vi $} & \multirow{3}{*}{Direct Inference} & None           & SFT        & Direct       & 51.09               & \multirow{3}{*}{50.91}   & \multirow{3}{*}{1}         & \multirow{3}{*}{1}         \\
                        &                                   & Pipeline-based & SFT        & Direct       & 51.29               &                          &                            &                            \\
                        &                                   & Bidirectional  & SFT        & Direct       & 50.36               &                          &                            &                            \\  \cmidrule(l){2-9}
                        & \multirow{2}{*}{Intersection Inference}     & Pipeline-based & SFT        & Intersection & 52.70               & \multirow{2}{*}{52.52}   & \multirow{2}{*}{10}        & \multirow{2}{*}{20}        \\
                        &                                   & Bidirectional  & SFT        & Intersection & 52.34               &                          &                            &                            \\ \cmidrule(l){2-9}
                        & \multirow{2}{*}{Union Inference}            & Pipeline-based & SFT        & Union        & 51.03               & \multirow{2}{*}{51.15}   & \multirow{2}{*}{2}         & \multirow{2}{*}{4}         \\
                        &                                   & Bidirectional  & SFT        & Union        & 51.27               &                          &                            &                            \\ \bottomrule
\end{tabular}}
\caption{Calculation example of the three variables based on Qwen3-8B in the CMeIE-V2. $vd, vt, vi \in \mathbb{Z}$.}
\label{tab:app_ac}
\end{table*}

\begin{table*}[!ht]
\centering
\scalebox{0.65}{
\begin{tabular}{p{0.4cm}<{\centering}p{2.4cm}p{2.2cm}p{2.0cm}<{\centering}p{1.2cm}<{\centering}p{0.8cm}<{\centering}p{0.3cm}<{\centering}p{0.3cm}<{\centering}p{0.3cm}<{\centering}p{1.1cm}<{\centering}p{0.8cm}<{\centering}p{0.3cm}<{\centering}p{0.3cm}<{\centering}p{0.3cm}<{\centering}p{1.1cm}<{\centering}p{0.8cm}<{\centering}p{0.3cm}<{\centering}p{0.3cm}<{\centering}p{0.3cm}<{\centering}}
\toprule
\multirow{3}{*}{\textbf{ID}} & \multicolumn{3}{c}{\textbf{Combination}}                                                        & \multicolumn{5}{c}{\textbf{GLM4-9B}}                                                    & \multicolumn{5}{c}{\textbf{InternLM3-8B}}                                               & \multicolumn{5}{c}{\textbf{LLaMA3.1-8B}}                                                \\  \cmidrule(l){2-4} \cmidrule(l){5-9} \cmidrule(l){10-14} \cmidrule(l){15-19}
                    & \multirow{2}{*}{Data}  & \multirow{2}{*}{Tuning}    & \multirow{2}{*}{Inference} & \multirow{2}{*}{Time(s)} & \multirow{2}{*}{F1(\%)} & \multicolumn{3}{c}{\textbf{Variable}} & \multirow{2}{*}{Time(s)} & \multirow{2}{*}{F1(\%)} & \multicolumn{3}{c}{\textbf{Variable}} & \multirow{2}{*}{Time(s)} & \multirow{2}{*}{F1(\%)} & \multicolumn{3}{c}{\textbf{Variable}} \\ \cmidrule(l){7-9} \cmidrule(l){12-14} \cmidrule(l){17-19}
                    &                        &                           &                            &                          &                     & $ vd $   & $ vt $   & $ vi $  &                          &                     & $ vd $   & $ vt $   & $ vi $  &                          &                     & $ vd $   & $ vt $   & $ vi $  \\ \midrule
1                   & \multirow{4}{*}{None}     & SFT                     & Direct                  & 10,774                                       & 49.99               & 1        & 6        & 1       & 9,725                    & 48.12               & 1        & 10       & 1       & 10,382                   & 51.06               & 7        & 10       & 2       \\
2                   &                         & SFT + GRPO                  & Direct                  & 17,208                                       & 50.59               & 1        & 10       & 1       & 22,067                   & 47.42               & 1        & 3        & 1       & 19,329                   & 50.56               & 7        & 8        & 2       \\
3                   &                         & SFT + DPO                  & Direct                  & 11,125                                       & 49.15               & 1        & 1        & 1       & 13,101                   & 47.50                & 1        & 4        & 1       & 12,231                   & 49.45               & 7        & 4        & 2       \\
4                   &                         & SFT + KTO                  & Direct                  & 15,231                                       & 50.35               & 1        & 8        & 1       & 16,842                   & 47.22               & 1        & 1        & 1       & 15,558                   & 48.70                & 7        & 1        & 2       \\  \cmidrule(l){2-19}
5                   & \multirow{13}{*}{Pipeline-based} & \multirow{3}{*}{SFT}    & Direct                  & 64,473                                       & 53.07               & 10       & 6        & 1       & 74,118                   & 49.87               & 10       & 10       & 1       & 74,032                   & 50.87               & 1        & 10       & 2       \\
6                   &                         &                        & Intersection                  & 65,493                                       & 53.55               & 10       & 6        & 10      & 75,173                   & 50.61               & 10       & 10       & 10      & 75,232                   & 51.92               & 1        & 10       & 10      \\
7                   &                         &                        & Union                  & 65,493                                       & 52.59               & 10       & 6        & 4       & 75,173                   & 50.77               & 10       & 10       & 9       & 75,232                   & 50.97               & 1        & 10       & 1       \\ \cmidrule(l){3-19}
8                   &                         & \multirow{3}{*}{SFT + GRPO} & Direct                  & 112,834                                      & 52.61               & 10       & 10       & 1       & 142,384                  & 49.77               & 10       & 3        & 1       & 125,733                  & 51.40                & 1        & 8        & 2       \\
9                   &                         &                        & Intersection                  & 113,898                                      & 53.08               & 10       & 10       & 10      & 143,456                  & 49.90                & 10       & 3        & 10      & 126,936                  & 52.05               & 1        & 8        & 10      \\
10                  &                         &                        & Union                  & 113,898                                      & 52.47               & 10       & 10       & 4       & 143,456                  & 50.42               & 10       & 3        & 9       & 126,936                  & 51.14               & 1        & 8        & 1       \\ \cmidrule(l){3-19}
11                  &                         & \multirow{3}{*}{SFT + DPO} & Direct                  & 76,282                                       & 52.18               & 10       & 1        & 1       & 94,263                   & 49.79               & 10       & 4        & 1       & 84,492                   & 50.45               & 1        & 4        & 2       \\
12                  &                         &                        & Intersection                  & 77,322                                       & 52.92               & 10       & 1        & 10      & 95,308                   & 50.07               & 10       & 4        & 10      & 85,692                   & 51.40                & 1        & 4        & 10      \\
13                  &                         &                        & Union                  & 77,322                                       & 51.90                & 10       & 1        & 4       & 95,308                   & 49.74               & 10       & 4        & 9       & 85,692                   & 50.09               & 1        & 4        & 1       \\ \cmidrule(l){3-19}
14                  &                         & \multirow{3}{*}{SFT + KTO} & Direct                  & 94,828                                       & 51.39               & 10       & 8        & 1       & 120,054                  & 49.23               & 10       & 1        & 1       & 101,845                  & 50.74               & 1        & 1        & 2       \\
15                  &                         &                        & Intersection                  & 95,914                                       & 51.89               & 10       & 8        & 10      & 121,099                  & 50.01               & 10       & 1        & 10      & 103,031                  & 51.31               & 1        & 1        & 10      \\
16                  &                         &                        & Union                  & 95,914                                       & 50.90                & 10       & 8        & 4       & 121,099                  & 49.59               & 10       & 1        & 9       & 103,031                  & 51.46               & 1        & 1        & 1       \\ \cmidrule(l){2-19}
17                  & \multirow{13}{*}{Bidirectional} & \multirow{3}{*}{SFT}    & Direct                  & 66,241                                       & 51.14               & 4        & 6        & 1       & 70,909                   & 49.15               & 6        & 10       & 1       & 70,388                   & 51.16               & 10       & 10       & 2       \\
18                  &                         &                        & Intersection                  & 67,725                                       & 53.08               & 4        & 6        & 10      & 72,430                   & 50.28               & 6        & 10       & 10      & 72,032                   & 52.30                & 10       & 10       & 10      \\
19                  &                         &                        & Union                  & 67,725                                       & 51.62               & 4        & 6        & 4       & 72,430                   & 49.73               & 6        & 10       & 9       & 72,032                   & 50.89               & 10       & 10       & 1       \\ \cmidrule(l){3-19}
20                  &                         & \multirow{3}{*}{SFT + GRPO} & Direct                  & 112,431                                      & 51.54               & 4        & 10       & 1       & 133,803                  & 49.28               & 6        & 3        & 1       & 122,680                  & 50.99               & 10       & 8        & 2       \\
21                  &                         &                        & Intersection                  & 113,889                                      & 53.19               & 4        & 10       & 10      & 135,323                  & 50.23               & 6        & 3        & 10      & 124,384                  & 52.62               & 10       & 8        & 10      \\
22                  &                         &                        & Union                  & 113,889                                      & 51.65               & 4        & 10       & 4       & 135,323                  & 50.48               & 6        & 3        & 9       & 124,384                  & 51.30                & 10       & 8        & 1       \\ \cmidrule(l){3-19}
23                  &                         & \multirow{3}{*}{SFT + DPO} & Direct                  & 77,128                                       & 51.24               & 4        & 1        & 1       & 90,983                   & 49.69               & 6        & 4        & 1       & 83,879                   & 50.76               & 10       & 4        & 2       \\
24                  &                         &                        & Intersection                  & 78,645                                       & 52.90                & 4        & 1        & 10      & 92,469                   & 50.44               & 6        & 4        & 10      & 85,545                   & 51.96               & 10       & 4        & 10      \\
25                  &                         &                        & Union                  & 78,645                                       & 51.35               & 4        & 1        & 4       & 92,469                   & 49.88               & 6        & 4        & 9       & 85,545                   & 50.52               & 10       & 4        & 1       \\ \cmidrule(l){3-19}
26                  &                         & \multirow{3}{*}{SFT + KTO} & Direct                  & 95,757                                       & 51.97               & 4        & 8        & 1       & 118,323                  & 48.47               & 6        & 1        & 1       & 106,654                  & 51.12               & 10       & 1        & 2       \\
27                  &                         &                        & Intersection                  & 97,176                                       & 54.03               & 4        & 8        & 10      & 119,865                  & 49.83               & 6        & 1        & 10      & 108,316                  & 51.70                & 10       & 1        & 10      \\
28                  &                         &                        & Union                  & 97,176                                       & 51.85               & 4        & 8        & 4       & 119,865                  & 48.15               & 6        & 1        & 9       & 108,316                  & 51.20                & 10       & 1        & 1      
\\ \bottomrule
\end{tabular}}
\caption{Fitted data of the DTI planner $ f(x)_{dataset} $ on the CMeIE-V2 across three base-LLMs. The range of the variables' mapping is from 1 to 10.}
\label{tab:app_A444}
\end{table*}

\begin{table*}[t]
\centering
\scalebox{0.75}{
\begin{tabular}{p{0.4cm}<{\centering}p{2.1cm}p{2.3cm}p{2.0cm}<{\centering}p{1.5cm}<{\centering}p{1.2cm}<{\centering}p{0.5cm}<{\centering}p{0.5cm}<{\centering}p{0.5cm}<{\centering}p{1.5cm}<{\centering}p{1.2cm}<{\centering}p{0.5cm}<{\centering}p{0.5cm}<{\centering}p{0.5cm}<{\centering}}
\toprule
\multirow{3}{*}{\textbf{ID}} & \multicolumn{3}{c}{\textbf{Combination}}                      & \multicolumn{5}{c}{\textbf{ACE05}}                                                      & \multicolumn{5}{c}{\textbf{14Lap}}                                                              \\  \cmidrule(l){2-4} \cmidrule(l){5-9} \cmidrule(l){10-14}
                    & \multirow{2}{*}{Data}  & \multirow{2}{*}{Tuning} & \multirow{2}{*}{Inference}                                            & \multirow{2}{*}{Time(s)} & \multirow{2}{*}{F1(\%)} & \multicolumn{3}{c}{\textbf{Variable}} & \multirow{2}{*}{Time(s)} & \multirow{2}{*}{F1(\%)} & \multicolumn{3}{c}{\textbf{Variable}}                \\   \cmidrule(l){7-9} \cmidrule(l){12-14}
                    &                    &                     &  &                          &                     & $ vd $   & $ vt $   & $ vi $  &                          &                      & $ vd $ & $ vt $ & \multicolumn{1}{c}{$ vi $} \\ \midrule
1                   & \multirow{4}{*}{None}     & SFT                     & Direct                  & 3,843                    & 46.06               & 1        & 10       & 1       & 1,423                    & 60.68               & 1        & 6        & 1       \\
2                   &                         & SFT + GRPO                  & Direct                  & 4,827                    & 45.74               & 1        & 7        & 1       & 1,889                    & 61.58               & 1        & 10       & 1       \\
3                   &                         & SFT + DPO                  & Direct                  & 3,742                    & 45.46               & 1        & 4        & 1       & 1,269                    & 59.58               & 1        & 1        & 1       \\
4                   &                         & SFT + KTO                  & Direct                  & 4,211                    & 45.19               & 1        & 1        & 1       & 1,491                    & 61.22               & 1        & 8        & 1       \\  \cmidrule(l){2-14}
5                   & \multirow{13}{*}{Pipeline-based} & \multirow{3}{*}{SFT}    & Direct                  & 18,921                   & 51.98               & 10       & 10       & 1       & 8,748                    & 61.52               & 9        & 6        & 1       \\
6                   &                         &                        & Intersection                  & 18,956                   & 52.05               & 10       & 10       & 10      & 8,769                    & 62.38               & 9        & 6        & 10      \\
7                   &                         &                        & Union                  & 18,956                   & 53.06               & 10       & 10       & 10      & 8,769                    & 62.59               & 9        & 6        & 5       \\  \cmidrule(l){3-14}
8                   &                         & \multirow{3}{*}{SFT + GRPO} & Direct                  & 20,989                   & 53.16               & 10       & 7        & 1       & 13,202                   & 61.71               & 9        & 10       & 1       \\
9                   &                         &                        & Intersection                  & 21,022                   & 54.27               & 10       & 7        & 10      & 13,221                   & 63.63               & 9        & 10       & 10      \\
10                  &                         &                        & Union                  & 21,022                   & 53.60                & 10       & 7        & 10      & 13,221                   & 60.85               & 9        & 10       & 5       \\  \cmidrule(l){3-14}
11                  &                         & \multirow{3}{*}{SFT + DPO} & Direct                  & 16,387                   & 52.86               & 10       & 4        & 1       & 9,274                    & 62.00                  & 9        & 1        & 1       \\
12                  &                         &                        & Intersection                  & 16,387                   & 53.74               & 10       & 4        & 10      & 9,294                    & 65.15               & 9        & 1        & 10      \\
13                  &                         &                        & Union                  & 16,387                   & 52.61               & 10       & 4        & 10      & 9,294                    & 62.49               & 9        & 1        & 5       \\ \cmidrule(l){3-14}
14                  &                         & \multirow{3}{*}{SFT + KTO} & Direct                  & 18,788                   & 52.56               & 10       & 1        & 1       & 10,257                   & 61.34               & 9        & 8        & 1       \\
 15             &              &                        & Intersection                  & 18,788                   & 54.10                & 10       & 1        & 10      & 10,276                   & 64.80                & 9        & 8        & 10      \\
16                  &                         &                        & Union                  & 18,788                   & 52.57               & 10       & 1        & 10      & 10,276                   & 61.98               & 9        & 8        & 5       \\  \cmidrule(l){2-14}
17                  & \multirow{13}{*}{Bidirectional} & \multirow{3}{*}{SFT}    & Direct                  & 16,387                   & 51.21               & 9        & 10       & 1       & 7,890                    & 61.63               & 10       & 6        & 1       \\
18                  &                         &                        & Intersection                  & 16,417                   & 53.87               & 9        & 10       & 10      & 7,909                    & 63.92               & 10       & 6        & 10      \\
19                  &                         &                        & Union                  & 16,417                   & 52.73               & 9        & 10       & 10      & 7,909                    & 61.74               & 10       & 6        & 5       \\ \cmidrule(l){3-14}
20                  &                         & \multirow{3}{*}{SFT + GRPO} & Direct                  & 21,354                   & 51.69               & 9        & 7        & 1       & 12,599                   & 62.42               & 10       & 10       & 1       \\
21                  &                         &                        & Intersection                  & 21,386                   & 54.88               & 9        & 7        & 10      & 12,616                   & 63.29               & 10       & 10       & 10      \\
22                  &                         &                        & Union                  & 21,386                   & 52.31               & 9        & 7        & 10      & 12,616                   & 61.45               & 10       & 10       & 5       \\ \cmidrule(l){3-14}
23                  &                         & \multirow{3}{*}{SFT + DPO} & Direct                  & 16,621                   & 52.52               & 9        & 4        & 1       & 8,694                    & 61.64               & 10       & 1        & 1       \\
24                  &                         &                        & Intersection                  & 16,621                   & 53.21               & 9        & 4        & 10      & 8,713                    & 64.88               & 10       & 1        & 10      \\
25                  &                         &                        & Union                  & 16,621                   & 51.93               & 9        & 4        & 10      & 8,713                    & 61.50                & 10       & 1        & 5       \\ \cmidrule(l){3-14}
26                  &                         & \multirow{3}{*}{SFT + KTO} & Direct                  & 19,245                   & 50.82               & 9        & 1        & 1       & 9,467                    & 62.18               & 10       & 8        & 1       \\
27                  &                         &                        & Intersection                  & 19,245                   & 52.75               & 9        & 1        & 10      & 9,490                    & 65.96               & 10       & 8        & 10      \\
28                  &                         &                        & Union                  & 19,245                   & 53.13               & 9        & 1        & 10      & 9,490                    & 60.73               & 10       & 8        & 5      
\\ \bottomrule
\end{tabular}}
\caption{Fitted data of the DTI planner $ f(x)_{model} $ on Qwen3-8B for two task datasets. The range of the variables' mapping is from 1 to 10.}
\label{tab:app2}
\end{table*}

\section{Calculation Method of the Three Variables}\label{app:C}

This section represents the calculation method of the three variables. Each variable corresponds to a different stage in the model workflow. Representative F1 results from the corresponding configurations are used as the values of each variable.
The specific principles for mapping DTI strategies to numerical variables are shown in Table~\ref{tab:app_AA2}.
\begin{itemize}
    \item The data decomposition variable ($v_d$) represents how the dataset is structured, including None (no decomposition), Pipeline-based decomposition, and Bidirectional decomposition. For None (no decomposition), it is F1 (None, SFT, Direct). For Pipeline-based decomposition, it is F1 (Pipeline-based, SFT, Direct). For Bidirectional decomposition, it is F1 (Bidirectional, SFT, Direct).
    \item The tuning variable ($v_t$) describes the instruction tuning method, including SFT, SFT + GRPO, SFT + DPO, and SFT + KTO. For SFT, it is F1 (None, SFT, Direct). For SFT + GRPO, it is F1 (None, SFT + GRPO, Direct). For SFT + DPO, it is F1 (None, SFT + DPO, Direct). For SFT + KTO, it is F1 (None, SFT + KTO, Direct).
    \item The inference variable ($v_i$) denotes the inference strategy. For Direct, the value is (F1 (None, SFT, Direct) + F1 (Pipeline-based, SFT, Direct) + F1 (Bidirectional, SFT, Direct)) / 3. For Intersection, it is (F1 (Pipeline-based, SFT, Intersection) + F1 (Bidirectional, SFT, Intersection)) / 2, and for Union, it is (F1 (Pipeline-based, SFT, Union) + F1 (Bidirectional, SFT, Union)) / 2.
\end{itemize}

An example of the calculation for the three variables based on Qwen3-8B on the CMeIE-V2 dataset is illustrated in Table~\ref{tab:app_ac}.
Based on the value assignment rules in Table~\ref{tab:app_AA2}, the detailed process of calculating $vd$ is as follows:

\textit{(1) Obtaining F1 scores from Table~\ref{tab:4}.}

\begin{itemize}
    \item None: $F1 = 51.09$
    \item Pipeline-based: $F1 = 51.29$
    \item Bidirectional: $F1 = 50.36$
\end{itemize}

\textit{(2) Determining the mapping range.}

According to the rules, the value range is the integers from 1 to 10. Therefore, the method with the lowest F1 score is mapped to $1$, and the method with the highest F1 score is mapped to $10$:

\begin{itemize}
    \item $vd_{Bidirectional} = 1$
    \item $vd_{Pipeline-based} = 10$
\end{itemize}

\textit{(3) Computing the mapped value.}

Using linear proportional mapping:
\[
vd_{None} 
= \frac{51.09 - 50.36}{(51.29 - 50.36)/9} + 1.
\]
After rounding to the nearest integer:
\[
vd_{None} \approx 8.
\]

$vt$ and $vi$'s calculation method is the same as $vd$'s.

\begin{table*}[htb]
\centering
\scalebox{0.78}{
\begin{tabular}{p{3.9cm}p{2.5cm}p{1.0cm}<{\centering}p{1.0cm}<{\centering}p{1.0cm}<{\centering}p{1.0cm}<{\centering}p{1.0cm}<{\centering}p{1.0cm}<{\centering}p{1.0cm}<{\centering}p{1.0cm}<{\centering}p{1.0cm}<{\centering}}
\toprule
\textbf{Objective}                  & \textbf{Planner}         & \textbf{a}     & \textbf{b}     & \textbf{c}    & \textbf{d}    & \textbf{e}    & \textbf{f}    & \textbf{g}     & \textbf{h}    & \textbf{i}    \\ \midrule
\multirow{4}{*}{Performance }        & $ f(x)_{dataset} $  
&-1.75	&-3.49	&-2.07	&2.11	&0.61	&0.14	&1.47	&39.26	&26.45   \\
 & $ f(x)_{model} $  &-6.30	&-10.00	&-7.70	&6.90	&8.00	&10.00	&-9.80	&0.50	&-0.20    \\	 \cmidrule(lr){2-11}						
& $ f(x)^{new}_{dataset} $   &-0.62	&-3.56	&-2.99	&1.89	&0.32	&3.36	&-10.34	&10.43	&19.90     \\& $ f(x)^{new}_{model} $&-5.23	&-3.79	&6.83	&1.75	&7.86	&6.76	&8.00	&-21.51	&0.65              \\ \bottomrule
\end{tabular}}
\caption{The coefficient of the DTI planner in the objective of performance, where the range of the variables' mapping is from 1 to 10. ``new'' denotes a new set of coefficients compared to the Table~\ref{tab:3}.}
\label{tab:app_A2}
\end{table*}

\begin{table*}[!bt]
\centering
\scalebox{0.7}{
\begin{tabular}{p{0.5cm}<{\centering}p{0.8cm}<{\centering}p{2.1cm}p{2.1cm}<{\centering}p{1.cm}<{\centering}p{1.cm}<{\centering}p{1.cm}<{\centering}p{0.3cm}<{\centering}p{0.3cm}<{\centering}p{0.3cm}<{\centering}p{1.5cm}<{\centering}p{1.5cm}<{\centering}p{1.5cm}<{\centering}p{1.5cm}<{\centering}}
\toprule
\multirow{2}{*}{\textbf{ID}} & \multicolumn{3}{c}{\textbf{Combination}}                                      &  \multirow{2}{*}{\textbf{P(\%)}} & \multirow{2}{*}{\textbf{R(\%)}} & \multirow{2}{*}{\textbf{F1(\%)}} & \multicolumn{3}{c}{\textbf{Variable}} & \multirow{2}{*}{$ f(x)_{dataset} $} & \multirow{2}{*}{$ f(x)_{model} $} & \multirow{2}{*}{$ f(x)^{new}_{dataset} $} & \multirow{2}{*}{$ f(x)^{new}_{model} $} \\ \cmidrule(lr){2-4} \cmidrule(lr){8-10}
                             & Data                             & Tuning                      & Inference                                      &                                 &                                 &                                  & $ vd $      & $ vt $     & $ vi $     &                                     &                                   &                                           &                                         \\ \midrule
1                            & \multirow{4}{*}{\rotatebox{90}{None}}            & SFT                         & Direct                               & 50.76                           & 51.41                           & 51.09                            & 8           & 10         & 1          & 143                                 & -769                              & -170                                      & -587                                    \\
2                            &                                  & SFT + GRPO                  & Direct                                  & 49.95                           & 50.36                           & 50.15                            & 8           & 1          & 1          & -18                                 & -370                              & -78                                       & -205                                    \\
3                            &                                  & SFT + DPO                   & Direct                                & 50.83                           & 50.70                           & 50.76                            & 8           & 7          & 1          & 152                                 & -456                              & -75                                       & -391                                    \\
4                            &                                  & SFT + KTO                   & Direct                                  & 51.66                           & 50.11                           & 50.87                            & 8           & 8          & 1          & 156                                 & -540                              & -99                                       & -449                   \\ \cmidrule(l){2-14} 
5                            & \multirow{12}{*}{\rotatebox{90}{Pipeline-based}} & \multirow{3}{*}{SFT}        & Direct                            & 51.34                           & 51.25                           & 51.29                            & 10          & 10         & 1          & 126                                 & -861                              & -174                                      & -708                                    \\
6                            &                                  &                             & Intersection                           & 63.30                           & 45.14                           & 52.70                            & 10          & 10         & 10         & 227                                 & -5                                & 40                                        & 1,289                                   \\
7                            &                                  &                             & Union                                  & 45.49                           & 58.10                           & 51.03                            & 10          & 10         & 2          & 154                                 & -704                              & -127                                      & -541          \\ \cmidrule(lr){3-14}
8                            &                                  & \multirow{3}{*}{SFT + GRPO} & Direct                                 & 52.34                           & 51.99                           & 52.17                            & 10          & 1          & 1          & -73                                 & -586                              & -116                                      & -358                                    \\
9                            &                                  &                             & Intersection                           & 63.58                           & 45.36                           & 52.94                            & 10          & 1          & 10         & 16                                  & -540                              & -174                                      & 1,092                                   \\
10                           &                                  &                             & Union                                & 46.13                           & 58.11                           & 51.43                            & 10          & 1          & 2          & -46                                 & -520                              & -99                                       & -251                                    \\ \cmidrule(lr){3-14}
11                           &                                  & \multirow{3}{*}{SFT + DPO}  & Direct                                 & 52.42                           & 50.89                           & 51.64                            & 10          & 7          & 1          & 123                                 & -589                              & -91                                       & -523                                    \\
12                           &                                  &                             & Intersection                            & 63.10                           & 46.10                           & 53.28                            & 10          & 7          & 10         & 220                                 & -4                                & 33                                        & 1,292                                   \\
13                           &                                  &                             & Union                              & 44.78                           & 59.83                           & 51.22                            & 10          & 7          & 2          & 150                                 & -463                              & -53                                       & -376                                    \\  \cmidrule(lr){3-14}
14                           &                                  &   & Direct                                & 51.78                           & 51.63                           & 51.70                            & 10          & 8          & 1          & 131                                 & -660                              & -112                                      & -577                                    \\
\rowcolor{lightgreen!85} 15         &                                  &                             & Intersection                         & \textbf{65.41}                  & 46.42                           & \textbf{54.30}                   & 10          & 8          & 10         & \textbf{229}                        & \textbf{16}                       & \textbf{42}                               & \textbf{1,299}                          \\
16                           &                                  &        \multirow{-3}{*}{SFT + KTO}                     & Union                                & 45.90                           & 59.32                           & 51.75                            & 10          & 8          & 2          & 158                                 & -523                              & -71                                       & -424                 \\ \cmidrule(l){2-14} 
17                           & \multirow{12}{*}{\rotatebox{90}{Bidirectional}}  & \multirow{3}{*}{SFT}        & Direct                                 & 50.39                           & 50.33                           & 50.36                            & 1           & 10         & 1          & 91                                  & -842                              & -193                                      & -491                                    \\
18                           &                                  &                             & Intersection                           & 60.61                           & 46.05                           & 52.34                            & 1           & 10         & 10         & 142                                 & -634                              & -5                                        & 870                                     \\
19                           &                                  &                             & Union                               & 45.59                           & 58.58                           & 51.27                            & 1           & 10         & 2          & 113                                 & -757                              & -148                                      & -394                                    \\ \cmidrule(lr){3-14}
20                           &                                  & \multirow{3}{*}{SFT + GRPO} & Direct                              & 51.11                           & 50.48                           & 50.79                            & 1           & 1          & 1          & 63                                  & -9                                & 18                                        & 1                                       \\
21                           &                                  &                             & Intersection                         & 60.95                           & 45.57                           & 52.15                            & 1           & 1          & 10         & 103                                 & -611                              & -65                                       & 815                                     \\
22                           &                                  &                             & Union                                & 45.45                           & 58.38                           & 51.11                            & 1           & 1          & 2          & 84                                  & -14                               & 33                                        & 37                    \\ \cmidrule(lr){3-14}
23                           &                                  & \multirow{3}{*}{SFT + DPO}  & Direct                          & 50.72                           & 50.22                           & 50.47                            & 1           & 7          & 1          & 144                                 & -384                              & -58                                       & -259                                    \\
24                           &                                  &                             & Intersection                          & 62.32                           & 43.96                           & 51.55                            & 1           & 7          & 10         & 192                                 & -446                              & 39                                        & 920                                     \\
25                           &                                  &                             & Union                              & 43.52                           & 58.81                           & 50.02                            & 1           & 7          & 2          & 166                                 & -330                              & -24                                       & -182           \\ \cmidrule(lr){3-14}
26                           &                                  & \multirow{3}{*}{SFT + KTO}  & Direct                                & 52.58                           & 50.75                           & 51.65                            & 1           & 8          & 1          & 133                                 & -517                              & -96                                       & -328                                    \\
27                           &                                  &                             & Intersection                            & 63.34                           & 44.90                           & 52.55                            & 1           & 8          & 10         & 182                                 & -489                              & 32                                        & 911                                     \\
28                           &                                  &                             & Union                           & 44.79                           & 60.82                           & 51.59                            & 1           & 8          & 2          & 155                                 & -452                              & -58                                       & -245                                    \\ \bottomrule
\end{tabular}
}
\caption{The prediction results of constructing the DTI planner on the Qwen3-8B in CMeIE-V2.  The range of the variables' mapping is from 1 to 10.}
\label{tab:FF}
\end{table*}

\section{Detailed Fitted Data of the DTI Planner on CMeIE-V2 Dataset}\label{app:D}
This section presents the results of three LLMs (i.e., GLM4-9B, InternLM3-8B, and LLaMA3.1-8B) on CMeIE-V2, as reported in Table~\ref{tab:app_A444}. These results are used to solve the coefficients of the DTI planner. Specifically,
1) The ``F1'' is used to determine the optimal combination for single-objective optimization; ``F1 and Time'' is used to further refine the optimal combination for multi-objective optimization.
2) For each DTI combination, the values of the variables \(vd\), \(vt\), and \(vi\)
are calculated and substituted into the inequality formula~\ref{eq}.
3) The inequality solving results from three base-LLMs are combined to determine the coefficients of the DTI planner $ f(x)_{dataset} $.

\section{Detailed Fitted Data of the DTI Planner on Qwen3-8B Model}\label{app:E}
This section presents the results of Qwen3-8B on two different task datasets, EE (i.e.,  ACE05) and ABSA (i.e., 14Lap), as listed in Table~\ref{tab:app2}. These results are used to solve the coefficients of the DTI planner. Specifically,
1) The ``F1'' is used to determine the optimal combination for single-objective optimization; ``F1 and Time'' is used to further refine the optimal combination for multi-objective optimization.
2) For each DTI combination, the values of the variables \(vd\), \(vt\), and \(vi\)
are calculated and substituted into the inequality formula~\ref{eq}.
3) The inequality solving results from both datasets are combined to determine the coefficients of the DTI planner $ f(x)_{model} $.

\section{Analyze the Non-uniqueness of the Coefficient Solutions}\label{app:FF}
As shown in Table~\ref{tab:app_A2}, under the same search range, we can still stably obtain the optimal DTI combination ID 15 (as shown in Table~\ref{tab:FF}) by changing the coefficient solution of another set of planners. This result indicates that the information conveyed by the scalar encoding is adequate and effective for the current task, and does not impose a limitation on the final performance. In other words, the coefficients of the DTI planner are not a unique solution.


\begin{table*}[t]
\centering
\scalebox{0.78}{
\begin{tabular}{p{3.4cm}p{2.2cm}p{1.0cm}<{\centering}p{1.0cm}<{\centering}p{1.0cm}<{\centering}p{1.0cm}<{\centering}p{1.0cm}<{\centering}p{1.0cm}<{\centering}p{1.0cm}<{\centering}p{1.0cm}<{\centering}p{1.0cm}<{\centering}}
\toprule
\textbf{Objective}                  & \textbf{Planner}         & \textbf{a}     & \textbf{b}     & \textbf{c}    & \textbf{d}    & \textbf{e}    & \textbf{f}    & \textbf{g}     & \textbf{h}    & \textbf{i}    \\ \midrule
\multirow{2}{*}{Performance }        & $ f(x)_{dataset} $                 & -0.30                           & -3.21                          & -1.51                          & 1.17                           & -0.42                          & 2.70                            & -3.44                          & 37.51                          & 10.13                          \\
                                       & $ f(x)_{model}$                   & 0.20                            & -7.64                          & 7.62                           & 2.77                           & -1.69                          & 10.00                             & -9.96                          & -9.93                          & 4.87                          
\\ \bottomrule
\end{tabular}}
\caption{The coefficient of the DTI planner, where the range of the variables' mapping is from 1 to 20.}
\label{tab:app_A3}
\end{table*}

\begin{table*}[!bt]
\centering
\scalebox{0.7}{
\begin{tabular}{p{0.5cm}<{\centering}p{2.4cm}p{2.2cm}p{2.2cm}<{\centering}p{1.3cm}<{\centering}p{1.1cm}<{\centering}p{1.1cm}<{\centering}p{1.1cm}<{\centering}p{0.4cm}<{\centering}p{0.4cm}<{\centering}p{0.4cm}<{\centering}p{1.6cm}<{\centering}p{1.6cm}<{\centering}}
\toprule
\multirow{2}{*}{\textbf{ID}} & \multicolumn{3}{c}{\textbf{Combination}}                                       & \multirow{2}{*}{\textbf{Time(s)}} & \multirow{2}{*}{\textbf{P(\%)}} & \multirow{2}{*}{\textbf{R(\%)}} & \multirow{2}{*}{\textbf{F1(\%)}} & \multicolumn{3}{c}{\textbf{Variable}} & \multirow{2}{*}{$ f(x)_{dataset} $} & \multirow{2}{*}{$ f(x)_{model} $} \\ \cmidrule(lr){2-4} \cmidrule(lr){9-11}
                    & Data                   & Tuning                     & Inference &                          &                      &                      &                     & $ vd $ & $ vt $ & $ vi $ &                                     &                                   \\ \midrule
  \rowcolor{gray!12} 1                                   &                          & SFT                      & Direct                                  & 11,405                                        & 50.76                                     & 51.41                                     & 51.09                                    & 16                                  & 20                                  & 1                                   & -235                                                     & -2291                                                  \\
2                                   &                          & SFT + GRPO                   & Direct                                  & 17,342                                        & 49.95                                     & 50.36                                     & 50.15                                    & 16                                  & 1                                   & 1                                   & -74                                                      & -86                                                    \\
3                                   &                          & SFT + DPO                   & Direct                                  & 11,167                                        & 50.83                                     & 50.70                                      & 50.76                                    & 16                                  & 13                                  & 1                                   & 94                                                       & -837                                                   \\
4                                   & \multirow{-4}{*}{None}     & SFT + KTO                   & Direct                                  & 13,931                                        & 51.66                                     & 50.11                                     & 50.87                                    & 16                                  & 16                                  & 1                                   & -9                                                       & -1368                                                  \\  \cmidrule(lr){2-13}
5                                   &                          &                         & Direct                                  & 68,227                                        & 51.34                                     & 51.25                                     & 51.29                                    & 20                                  & 20                                  & 1                                   & -200                                                     & -2087                                                  \\
6                                   &                          &                         & Intersection                                  & 70,029                                        & 63.30                                      & 45.14                                     & 52.70                                     & 20                                  & 20                                  & 20                                  & 256                                                      & 4204                                                   \\
7                                   &                          & \multirow{-3}{*}{SFT}    & Union                                  & 70,029                                        & 45.49                                     & 58.10                                      & 51.03                                    & 20                                  & 20                                  & 4                                   & -56                                                      & -1460                                                  \\
8                                   &                          &                         & Direct                                  & 111,470                                       & 52.34                                     & 51.99                                     & 52.17                                    & 20                                  & 1                                   & 1                                   & -128                                                     & -93                                                    \\
9                                   &                          &                         & Intersection                                  & 113,262                                       & 63.58                                     & 45.36                                     & 52.94                                    & 20                                  & 1                                   & 20                                  & -646                                                     & 2588                                                   \\
10                                  &                          & \multirow{-3}{*}{SFT + GRPO} & Union                                  & 113,262                                       & 46.13                                     & 58.11                                     & 51.43                                    & 20                                  & 1                                   & 4                                   & -138                                                     & -35                                                    \\
11                                  &                          &                         & Direct                                  & 77,641                                        & 52.42                                     & 50.89                                     & 51.64                                    & 20                                  & 13                                  & 1                                   & 96                                                       & -711                                                   \\
12                                  &                          &                         & Intersection                                  & 79,593                                        & 63.10                                      & 46.10                                      & 53.28                                    & 20                                  & 13                                  & 20                                  & 193                                                      & 4250                                                   \\
13                                  &                          & \multirow{-3}{*}{SFT + DPO} & Union                                  & 79,593                                        & 44.78                                     & 59.83                                     & 51.22                                    & 20                                  & 13                                  & 4                                   & 184                                                      & -293                                                   \\
14                                  &                          &                         & Direct                                  & 94,879                                        & 51.78                                     & 51.63                                     & 51.70                                     & 20                                  & 16                                  & 1                                   & 7                                                        & -1209                                                  \\
\rowcolor{lightgreen!85} 15 &                          &                         & {Intersection} & {96,593}       & \textbf{65.41}    &{46.42}    & \textbf{54.30}    & {20} & {16} & {20} & \textbf{259}                     &\textbf{4322}                  \\
16                                  & \multirow{-12}{*}{Pipeline-based} & \multirow{-3}{*}{SFT + KTO} & Union                                  & 96,593                                        & 45.90                                      & 59.32                                     & 51.75                                    & 20                                  & 16                                  & 4                                   & 120                                                      & -701                                                   \\  \cmidrule(lr){2-13}
17                                  &                          &                         & Direct                                  & 65,185                                        & 50.39                                     & 50.33                                     & 50.36                                    & 1                                   & 20                                  & 1                                   & -452                                                     & -2998                                                  \\
18                                  &                          &                         & Intersection                                  & 66,804                                        & 60.61                                     & 46.05                                     & 52.34                                    & 1                                   & 20                                  & 20                                  & 156                                                      & 3903                                                   \\
19                                  &                          & \multirow{-3}{*}{SFT}    & Union                                  & 66,804                                        & 45.59                                     & 58.58                                     & 51.27                                    & 1                                   & 20                                  & 4                                   & -283                                                     & -2274                                                  \\
20                                  &                          &                         & Direct                                  & 110,836                                       & 51.11                                     & 50.48                                     & 50.79                                    & 1                                   & 1                                   & 1                                   & 43                                                       & -4                                                     \\
21                                  &                          &                         & Intersection                                  & 112,490                                       & 60.95                                     & 45.57                                     & 52.15                                    & 1                                   & 1                                   & 20                                  & -324                                                     & 3287                                                   \\
22                                  &                          & \multirow{-3}{*}{SFT + GRPO} & Union                                  & 112,490                                       & 45.45                                     & 58.38                                     & 51.11                                    & 1                                   & 1                                   & 4                                   & 57                                                       & 150                                                    \\
23                                  &                          &                         & Direct                                  & 76,539                                        & 50.72                                     & 50.22                                     & 50.47                                    & 1                                   & 13                                  & 1                                   & 0                                                        & -1253                                                  \\
24                                  &                          &                         & Intersection                                  & 78,212                                        & 62.32                                     & 43.96                                     & 51.55                                    & 1                                   & 13                                  & 20                                  & 249                                                      & 4318                                                   \\
25                                  &                          & \multirow{-3}{*}{SFT + DPO} & Union                                  & 78,212                                        & 43.52                                     & 58.81                                     & 50.02                                    & 1                                   & 13                                  & 4                                   & 112                                                      & -739                                                   \\
26                                  &                          &                         & Direct                                  & 88,759                                        & 52.58                                     & 50.75                                     & 51.65                                    & 1                                   & 16                                  & 1                                   & -155                                                     & -1909                                                  \\
27                                  &                          &                         & Intersection                                  & 90,352                                        & 63.34                                     & 44.90                                      & 52.55                                    & 1                                   & 16                                  & 20                                  & 248                                                      & 4231                                                   \\
28                                  & \multirow{-12}{*}{Bidirectional} & \multirow{-3}{*}{SFT + KTO} & Union                                  & 90,352                                        & 44.79                                     & 60.82                                     & 51.59                                    & 1                                   & 16                                  & 4                                   & -19                                                      & -1306                     
\\ \bottomrule
\end{tabular}}
\caption{The prediction results of constructing the DTI planner on the Qwen3-8B in CMeIE-V2, where the range of the variables' mapping is from 1 to 20.}
\label{tab:app_A4}
\end{table*}

\section{Analyze the Values of the Variables' Mapping Range}\label{app:F}
This section represents the analysis of the values of the variable mapping range. To further examine the robustness of the proposed PlanE framework with respect to the variable range, we extend the value range of the three variables from 1–10 to 1–20 and repeat all experiments under the same settings. The extended experimental results are presented in Table~\ref{tab:app_A3} and Table~\ref{tab:app_A4}, corresponding to Table~\ref{tab:3} and Table~\ref{tab:4}, respectively. The comparison shows that the optimal DTI configurations and overall performance trends remain consistent across both ranges. In both the 1–10 and 1–20 variable ranges, the validation phase consistently identifies the same optimal configuration (ID = 15). This indicates that enlarging the variable range does not affect the experimental outcomes, demonstrating the stability and reliability of PlanE in capturing the underlying relationships between data decomposition, tuning method, and inference strategy.

\begin{figure*}[t]     
\centering
\includegraphics[width=0.9\textwidth]{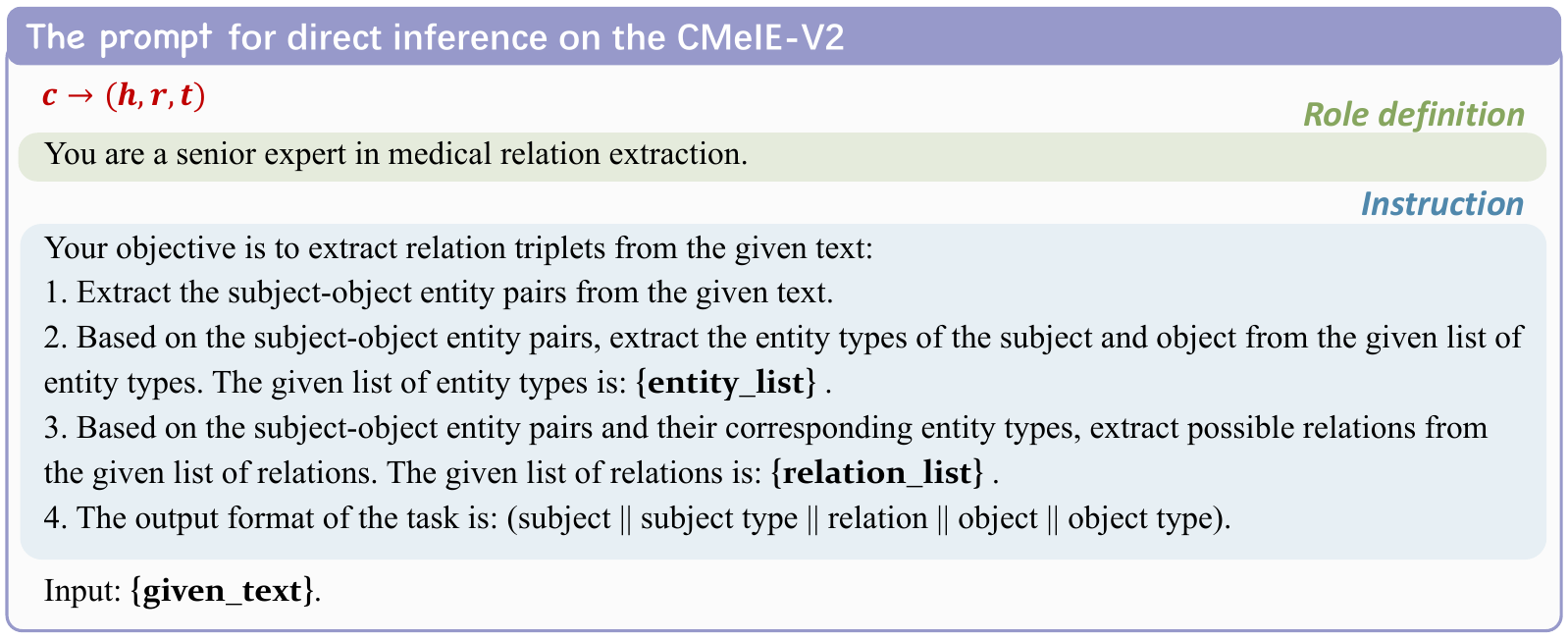}
\caption{The prompt for direct inference on the CMeIE-V2.}
\label{Fig.A1}
\end{figure*}

\begin{figure*}[!tb]     
\centering
\includegraphics[width=0.9\textwidth]{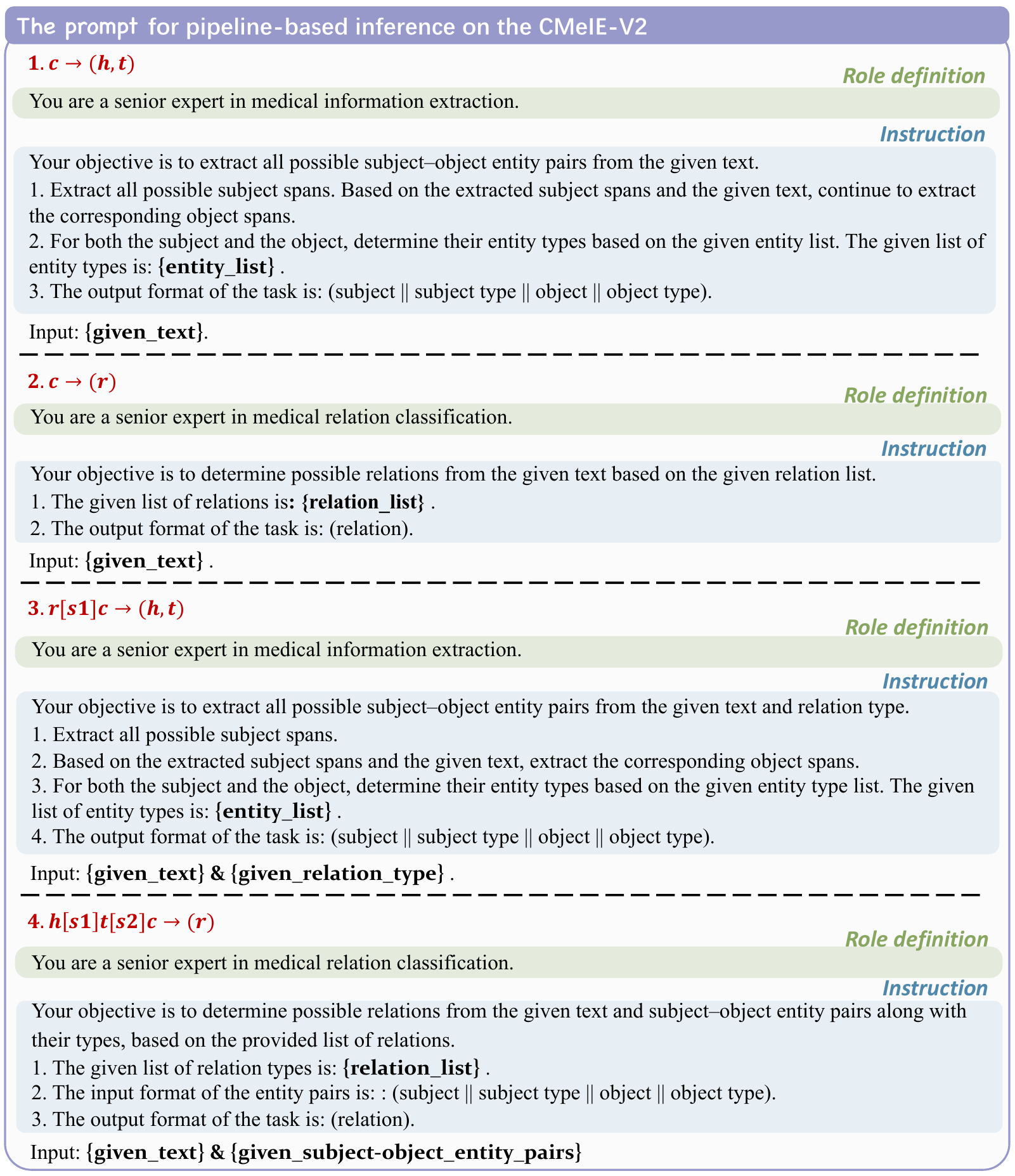}
\caption{The prompt for pipeline-based inference on the CMeIE-V2.}
\label{Fig.A2}
\end{figure*}

\section{Different Reinforcement Learning Methods}\label{app:H}
This section represents the three reinforcement learning methods used in this paper.

\textbf{GRPO}~\cite{shao2024deepseekmath}: This method optimizes the model based on group-relative rewards instead of pairwise comparisons. The core idea is to jointly evaluate a batch of model responses and use a static scoring function to assess the quality of each output from two perspectives — format-based and value-based rewards. The policy is then updated relative to the mean performance within the group, thereby improving training stability and convergence efficiency.
\begin{equation}
\begin{aligned}
\mathcal{J}_{\text{GRPO}}(\theta) = \mathbb{E} & \left[ q \sim P(Q), \{o_i\}_{i=1}^G \sim \pi_{\theta_{\text{old}}} \right] \\
\frac{1}{G} \sum_{i=1}^G & \left( \min \left( \frac{\pi_{\theta}(o_i|q)}{\pi_{\theta_{\text{old}}}(o_i|q)} A_i, \right. \right. \\
& \text{clip} \left( \frac{\pi_{\theta}(o_i|q)}{\pi_{\theta_{\text{old}}}(o_i|q)}, 1-\epsilon, 1+\epsilon \right) A_i \\
& \left. \vphantom{\frac{\pi_{\theta}}{\pi_{\theta}}} - \beta D_{KL} (\pi_{\theta} \| \pi_{\text{ref}}) \right),
\end{aligned}
\end{equation}
where \(\mathbb{E} \left[ q \sim P(Q), \{o_i\}_{i=1}^G \sim \pi_{\theta_{\text{old}}}(O|q) \right]\) 
denotes the expectation where the question \(q\) is sampled from the question distribution \(P(Q)\), and the output group \(\{o_i\}_{i=1}^G\) consists of \(G\) candidate responses sampled from the old policy \(\pi_{\theta_{\text{old}}}\). \(G\) represents the number of responses sampled for each question, \(\pi_{\theta}(o_i|q)\) denotes the probability of generating token \(o_i\) given question \(q\) under the current policy \(\pi_{\theta}\), and \(\pi_{\theta_{\text{old}}}(o_i|q)\) denotes the probability of generating token \(o_i\) given question \(q\) under the old policy \(\pi_{\theta_{\text{old}}}\). \(\epsilon\) is a clipping hyperparameter that limits the magnitude of policy updates, \(\beta\) represents the coefficient of the KL divergence penalty term, \(\pi_{\text{ref}}\) is the reference model, and \(D_{KL}\) is the KL regularization term that constrains the policy from deviating excessively from the reference model. Finally, \({A}_i\) denotes the advantage estimate of the \(i\)-th output, computed based on the group-relative rewards.

\textbf{DPO}~\cite{rafailov2023direct}: This method simplifies RLHF by eliminating the need for a reward model. Instead, it directly optimizes the policy toward preferred responses based on preference pairs. The core idea is to align the model distribution $\pi_\theta$ with the reference policy $\pi_{\text{ref}}$ using a logistic objective that favors higher-probability preferred outputs.

\begin{equation}
\begin{aligned}
\mathcal{L}_{\text{DPO}} & (\pi_\theta; \pi_{\text{ref}}) = -\mathbb{E}_{(x, y_w, y_l) \sim \mathcal{D}} \Bigl[ \log \sigma \Bigl( \\
& \beta \log \frac{\pi_\theta(y_w|x)}{\pi_{\text{ref}}(y_w|x)} - \beta \log \frac{\pi_\theta(y_l|x)}{\pi_{\text{ref}}(y_l|x)} \Bigr) \Bigr],
\end{aligned}
\end{equation}

where \(\pi_\theta\) denotes the trainable policy model, and \(\pi_{\text{ref}}\) represents the reference model that serves as a baseline policy.  
\(\mathcal{D}\) denotes the dataset of human preference pairs \((x, y_w, y_l)\),  
where \(x\) is the given input (e.g., a prompt or question), \(y_w\) is the preferred (winning) response, and \(y_l\) is the less-preferred (losing) response.  
\(\beta\) is a temperature coefficient controlling the sharpness of the preference contrast.  
\(\sigma(\cdot)\) represents the sigmoid function, converting the logit difference into a probability.

\textbf{KTO}~\cite{ethayarajh2024kto}: This method is inspired by behavioral economics, particularly \textit{Kahneman--Tversky Prospect Theory}, introducing asymmetric weighting for positive and negative feedback. The core idea is to encourage the model to be risk-averse for negative outcomes and risk-seeking for positive outcomes, improving controllability and alignment stability.
\begin{equation}
\mathcal{L}_{\text{KTO}}(\pi_\theta, \pi_{\text{ref}}) = 
\mathbb{E}_{x, y \sim \mathcal{D}} \left[ \lambda_y - v(x, y) \right],
\end{equation}

\begin{equation}
v(x, y) \!=\!
\begin{cases}
\lambda_D \sigma(\beta (r_\theta(x, y) \!-\! z_0)),&\text{if } y \sim y_{\text{d}}|x \\
\lambda_U \sigma(\beta (z_0 \!-\! r_\theta(x, y))),&\text{if } y \sim y_{\text{u}}|x
\end{cases}
\end{equation}
where \(\lambda_D\) and \(\lambda_U\) are hyperparameters that weight desirable and undesirable outputs, respectively;  
\(\sigma(\cdot)\) denotes the sigmoid function;  
\(\beta\) is the temperature coefficient controlling the scaling of the reward difference;  
\(r_\theta(x, y)\) is the reward function, and \(z_0\) is the KL divergence between the two policies.  
\(y_{\text{d}}\) and \(y_{\text{u}}\) indicate the desirable and undesirable outputs for a given input \(x\), respectively.



\section{Prompt Examples of Inference}\label{app:G}
\begin{figure*}[!bt]     
\centering
\includegraphics[width=0.9\textwidth]{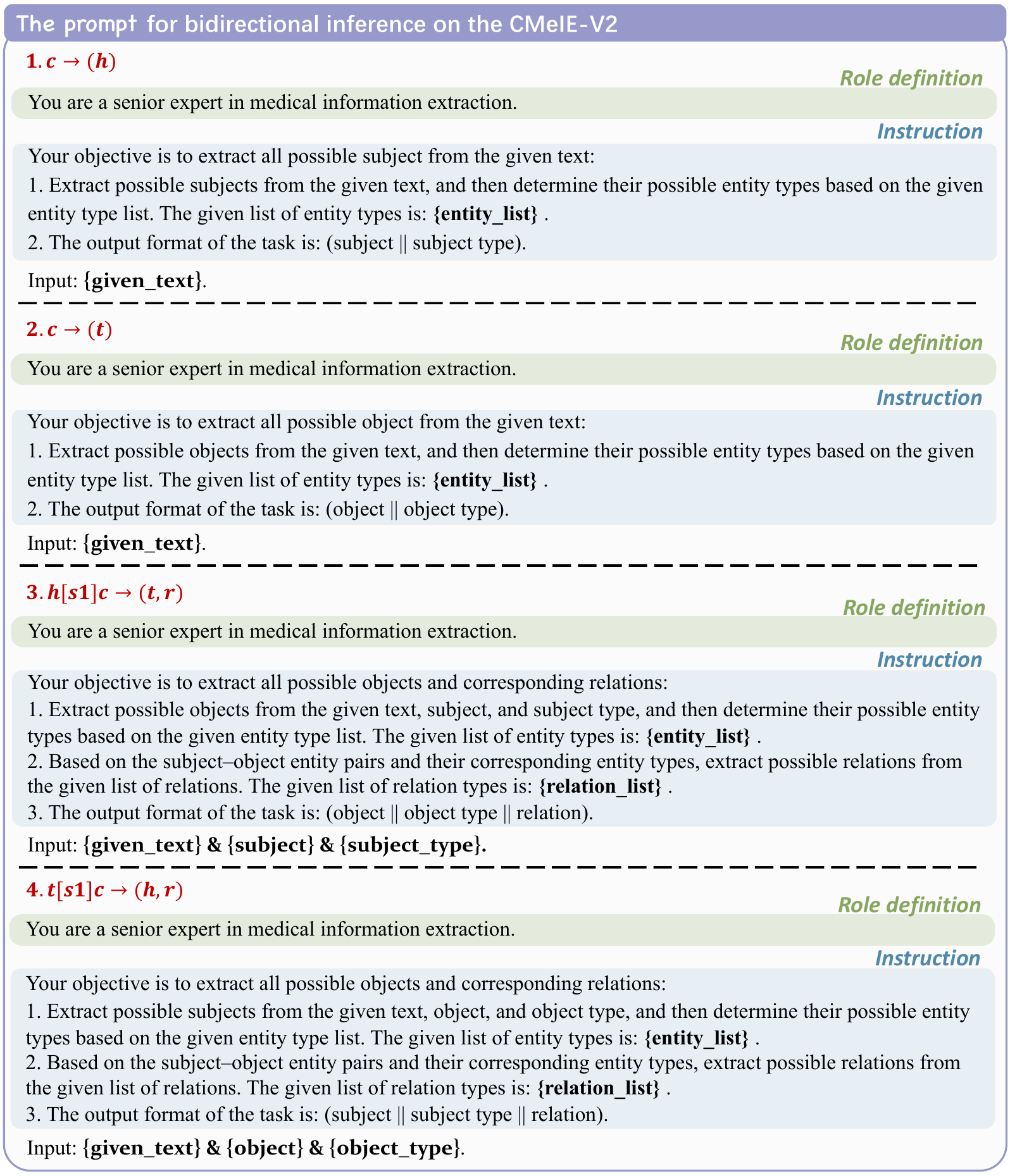}
\caption{The prompt for bidirectional inference on the CMeIE-V2.}
\label{Fig.A3}
\end{figure*}

\begin{figure*}[!bt]     
\centering
\includegraphics[width=0.9\textwidth]{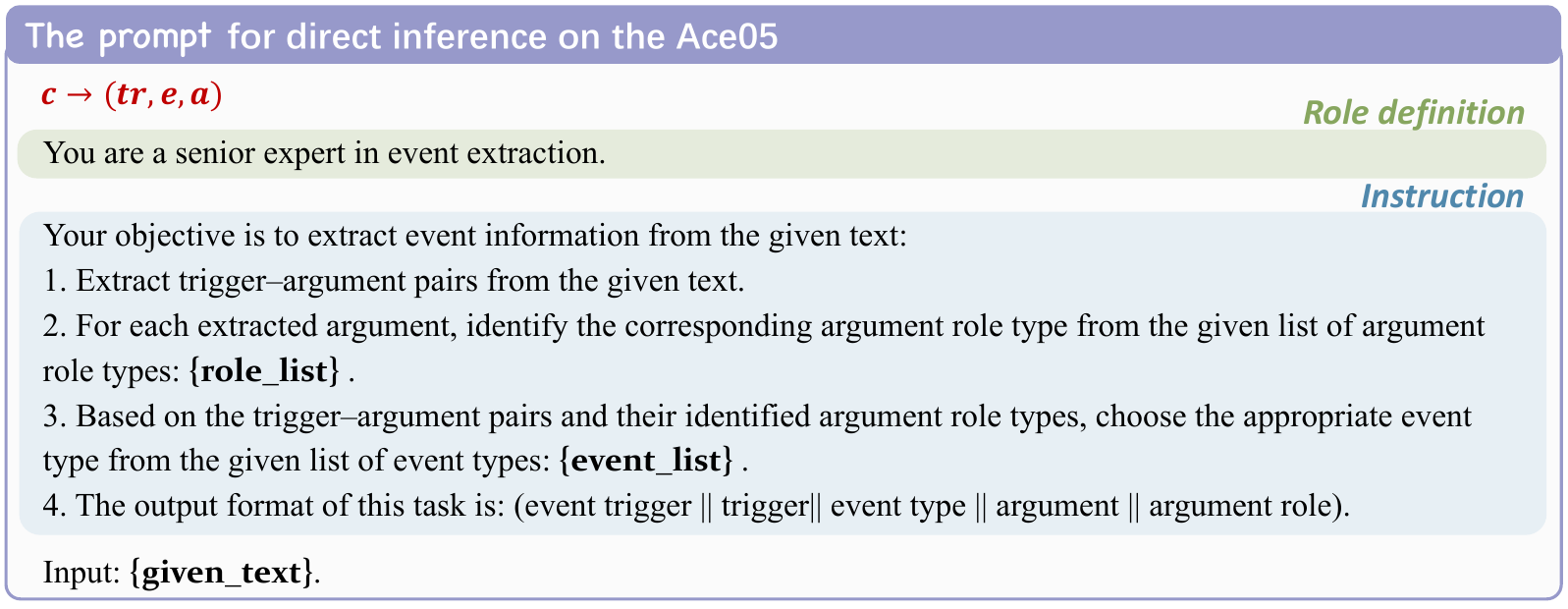}
\caption{The prompt for direct inference on the ACE05.}
\label{Fig.A4}
\end{figure*}

\begin{figure*}[!bt]     
\centering
\includegraphics[width=0.9\textwidth]{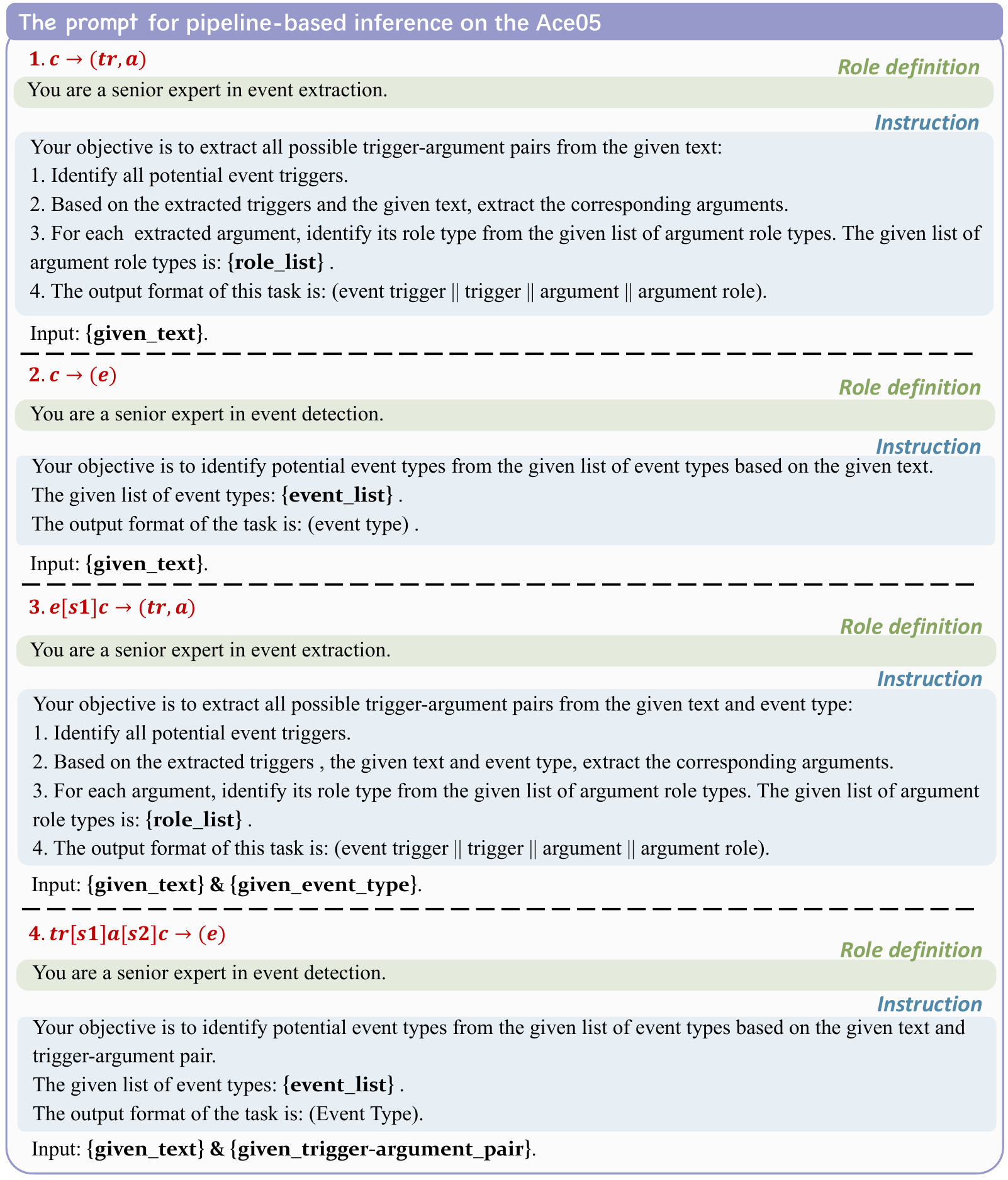}
\caption{The prompt for pipeline-based inference on the ACE05.}
\label{Fig.A5}
\end{figure*}

\begin{figure*}[!bt]     
\centering
\includegraphics[width=0.9\textwidth]{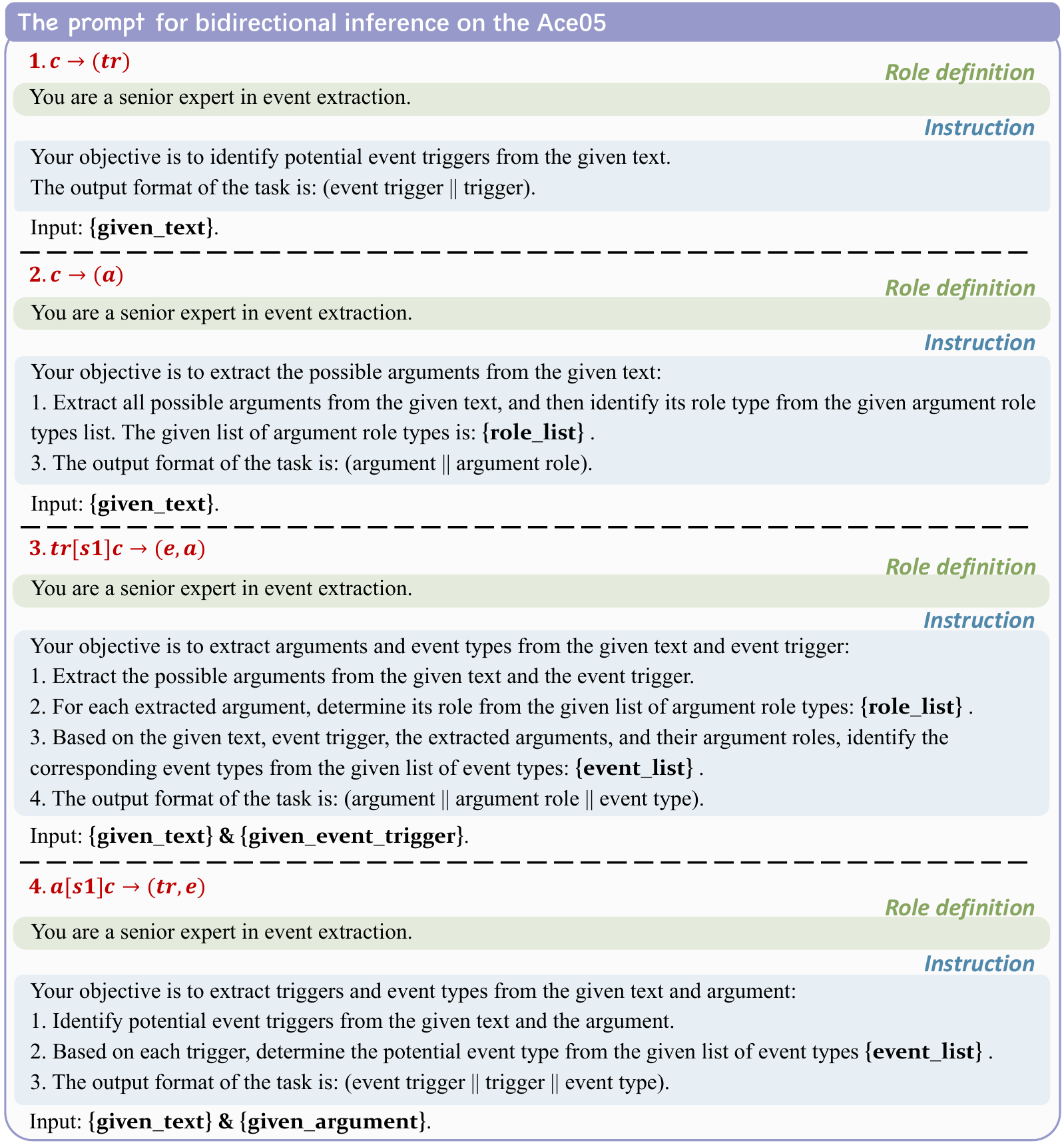}
\caption{The prompt for bidirectional inference on the ACE05.}
\label{Fig.A6}
\end{figure*}

\begin{figure*}[!bt]     
\centering
\includegraphics[width=0.9\textwidth]{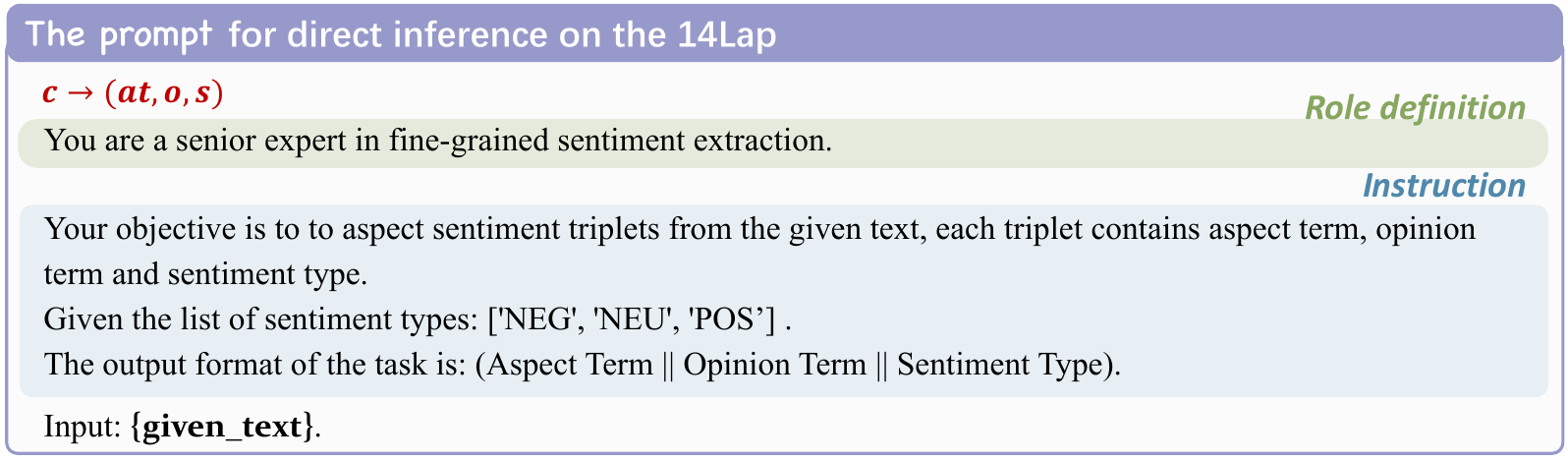}
\caption{The prompt for direct inference on the 14Lap.}
\label{Fig.A7}
\end{figure*}

\begin{figure*}[!bt]     
\centering
\includegraphics[width=0.9\textwidth]{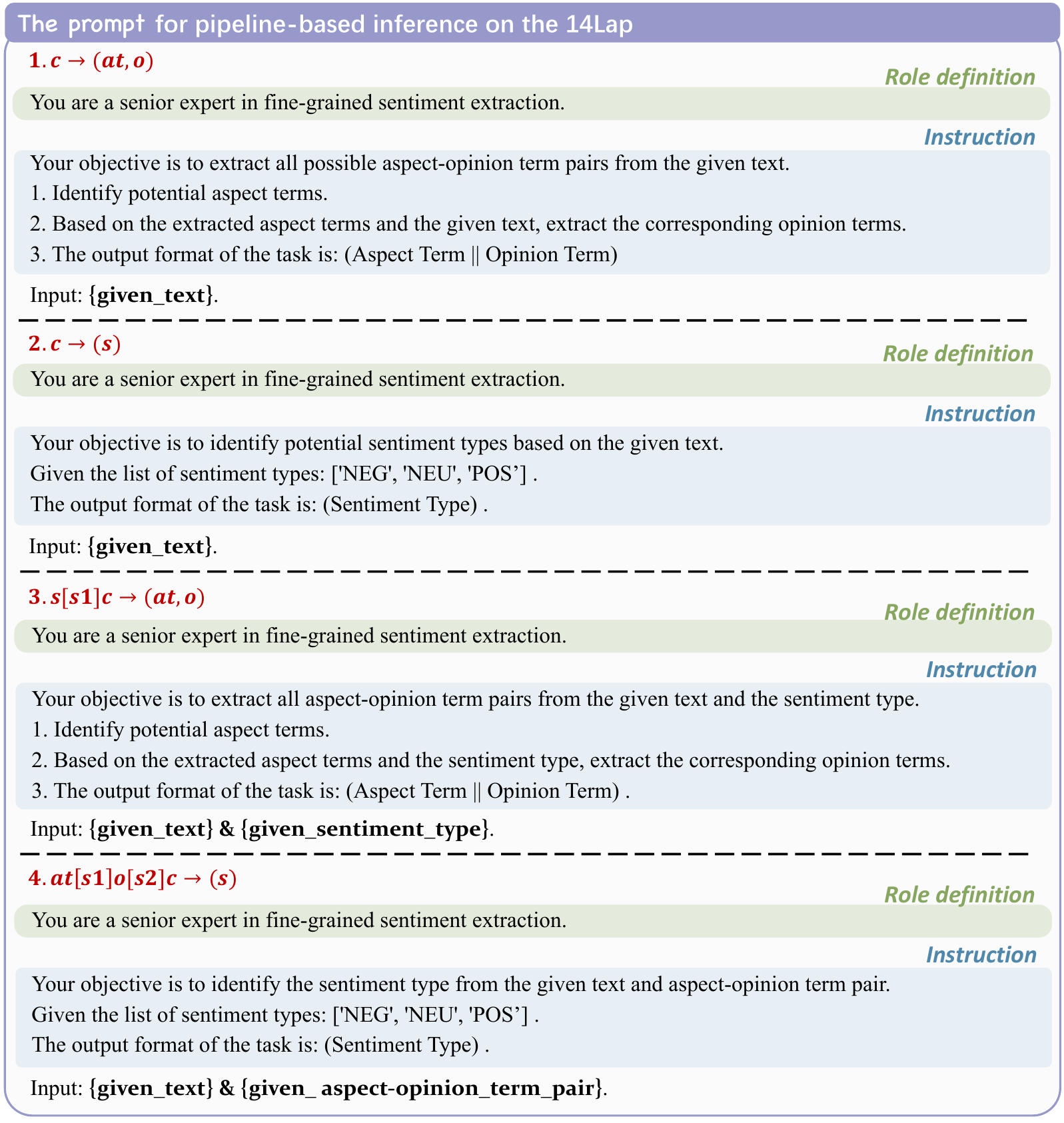}
\caption{The prompt for pipeline-based inference on the 14Lap.}
\label{Fig.A8}
\end{figure*}

\begin{figure*}[!bt]     
\centering
\includegraphics[width=0.9\textwidth]{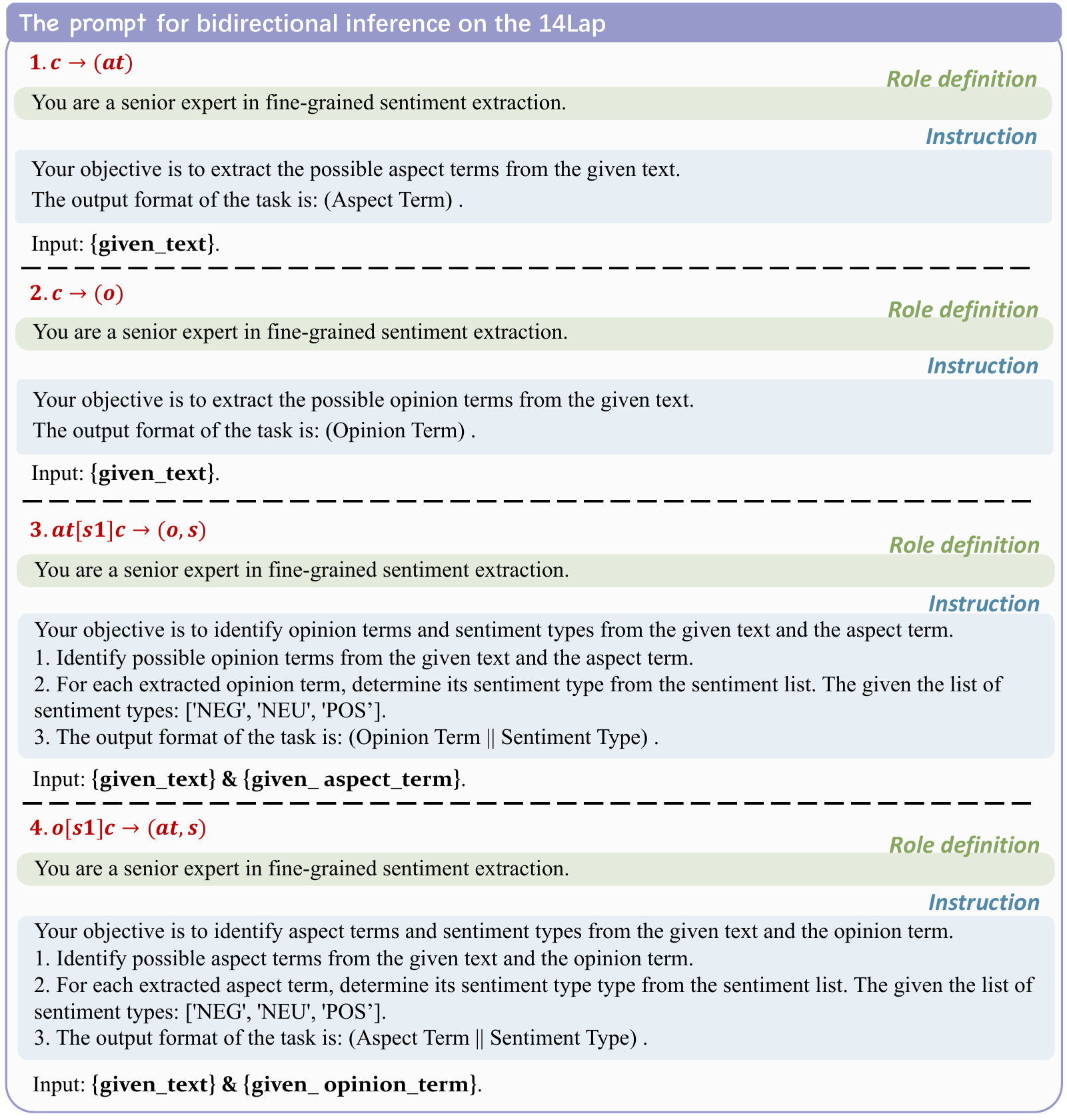}
\caption{The prompt for bidirectional inference on the 14Lap.}
\label{Fig.A9}
\end{figure*}

This section represents the prompt examples of inference on three datasets. Specifically, the prompt for direct inference on CMeIE-V2 is shown in Figure~\ref{Fig.A1}, the pipeline-based inference prompt in Figure~\ref{Fig.A2}, and the bidirectional inference prompt in Figure~\ref{Fig.A3}. For the ACE05 dataset, the prompts for direct, pipeline-based, and bidirectional inference are shown in Figures~\ref{Fig.A4},~\ref{Fig.A5}, and~\ref{Fig.A6}, respectively. Similarly, for the 14Lap dataset, the prompts for direct, pipeline-based, and bidirectional inference are presented in Figures~\ref{Fig.A7},~\ref{Fig.A8}, and~\ref{Fig.A9}, respectively.

\end{document}